%% file: main.tex
\UseRawInputEncoding
\documentclass[journal]{IEEEtran}
\IEEEoverridecommandlockouts
\usepackage{etoolbox}
\usepackage{cite}
\usepackage{amsmath,amssymb,amsfonts}
\usepackage{algorithmic}
\usepackage{graphicx}
\usepackage{textcomp}
\usepackage[OT1]{fontenc}
\usepackage{pifont}
\usepackage{multirow}
\usepackage{threeparttable}
\usepackage{booktabs}

\usepackage{hyperref}
\usepackage{caption}
\usepackage{tcolorbox}
\tcbuselibrary{listings, skins, breakable}
\usepackage{inconsolata}
\usepackage{lipsum}
\usepackage{float}
\usepackage{etoolbox}
\usepackage{siunitx}

\newtcolorbox{promptbox}[1]{%
  colback=gray!5,
  colframe=gray!60,
  coltitle=black,
  arc=3pt,
  sharp corners=south,
  boxrule=0.5pt,
  fonttitle=\bfseries,
  fontupper=\ttfamily\small,
  breakable,
  title={#1}
}

\lstdefinestyle{coltxt}{
  basicstyle=\ttfamily\footnotesize,
  breaklines=true, breakatwhitespace=true,
  columns=fullflexible,
  xleftmargin=0pt, framexleftmargin=0pt,
  frame=single, backgroundcolor=\color{gray!8},
  linewidth=\columnwidth, captionpos=b,
}

\usepackage{tabularx}
\usepackage[table,dvipsnames,svgnames]{xcolor}
\usepackage{booktabs}
\usepackage{array}
\usepackage{caption}
\usepackage{lscape} 
\renewcommand{\arraystretch}{1.3}
\definecolor{lightgray}{gray}{0.95}
\definecolor{headerblue}{RGB}{220,230,241}

\usepackage[dvipsnames,svgnames]{xcolor}
\usepackage{colortbl}

\definecolor{LightCyan}{rgb}{0.88,1,1}
\def\BibTeX{{\rm B\kern-.05em{\sc i\kern-.025em b}\kern-.08em
    T\kern-.1667em\lower.7ex\hbox{E}\kern-.125emX}}
\begin{document}
\bstctlcite{IEEEexample:BSTcontrol}
\title{Auto-DSM Under the Lens: \\ A Black-Box Evaluation Framework \\ for LLM-Based DSM Generation\\
}

\author{N.~Potters, T.~Hofman 
\thanks{N. Potters and T. Hofman (e-mail: t.hofman@tue.nl) are with the Eindhoven University of Technology (TU/e), Dept. of Mechanical Engineering, \href{https://www.tue.nl/en/research/research-groups/control-systems-technology}{Control Systems Technology} section, \href{https://www.tue.nl/en/research/research-groups/group-hofman}{Engineering Systems Design} group, P.O.Box 513, 5600 MB Eindhoven, The Netherlands.}}

\maketitle
\begin{abstract}
This paper presents a black-box evaluation framework to systematically assess the ability of Large Language Models (LLMs) to generate Design Structure Matrices (DSMs) from structured technical documentation. Motivated by the closed-source nature of current Auto-DSM pipelines, the framework introduces a reproducible methodology that benchmarks generated DSMs (GEN-DSMs) against manually validated ground-truth matrices (GT-DSMs). The evaluation integrates both single-run and multi-run perspectives, combining structural metrics (Completeness, Correctness, Coupling Density), classification metrics (Selective Accuracy, Abstention Coverage), and stability measures (Entropy, Fleiss' $\kappa$). To synthesize these aspects, a Composite Quality Score (Q) is proposed. Controlled experiments are conducted on two datasets: a fictive abstract system and a real-world refrigerator decomposition, covering variations in phrasing, parameter–dataset alignment, and system complexity. Results show that LLMs can produce structurally plausible DSMs and achieve high reproducibility under well-structured inputs, but remain sensitive to ambiguity, inconsistent dependency definitions, and prompt formulation. The findings highlight systematic sources of hallucination and abstention failure, demonstrating both the potential and current limitations of LLM-driven DSM automation. The proposed framework provides a transparent benchmark for auditing Auto-DSM pipelines and establishes foundations for integrating LLM-based decomposition methods into model-based systems engineering (MBSE) workflows.

\end{abstract}


\input{1_Introduction/Introduction}
\input{2_Background_and_Problem_Definition/Background_and_Problem_Definition}
\input{3_Methodology/Methodology}

\input{4_Experimental_Setup/Experimental_Setup}

\input{5_Results_Analysis/Results_Analysis}
\input{6_Discussion/Discussion}

\input{7_Conclusion_Future_work/Conclusion}

\section*{Acknowledgement}
The authors acknowledges that ChatGPT (powered by OpenAI's language model, GPT-4o; http://openai.com) was used for the purpose of grammatical check and content rephrasing. The editing was performed by the human author.

\bibliographystyle{IEEEtran}
\bibliography{IEEEabrv,IEEEReferences}

\input{A_Appendices/Appendix_A}
\end{document}

%% file: 1_Introduction/Introduction.tex
\section{Introduction}
The growing complexity of engineered systems has driven demand for tools and methods that support system-level analysis and decision-making processes. Among these, the Design Structure Matrix (DSM), introduced by Steward \cite{Steward1981}, remains a foundational method for visualizing and analyzing interdependencies between system elements. Its compact $N \times N$ format supports modularity analysis, design optimization, and process structuring across domains such as product development and systems engineering \cite{eppinger&browning2012}. Over decades, its use has matured through improvements in visualization and analysis tools—e.g., hierarchical clustering and dependency algorithms—forming a robust framework for downstream decision support \cite{eppinger&browning2012,browning2015}. 

The DSM-building phase—identifying components and interactions—remains manual, resource-intensive, and error-prone \cite{browning2015, eppinger&browning2012, clarkson2012, wilschut2018}. System decompositions typically rely on expert interviews and informal documentation, introducing variability, inconsistent terminology, and interpretation errors \cite{eppinger&browning2012, marwick2001, rus2002}. Poor documentation \cite{rubens2002} and tacit organizational knowledge \cite{rus2002} further undermine reproducibility. As noted in \cite{dong2002}, interviews are “just as important as reading design documents,” but they trade speed for accuracy, making DSM system decomposition costly and inconsistent.

Model-Based Systems Engineering (MBSE) provides a structured approach to managing system complexity and mitigating these issues \cite{madni2018}. By formalizing specifications in digital models, MBSE enables scalable and consistent dependency mapping \cite{DeSaquiSannes2022, Campo2023, Wilking2024}. A recent extension, the Elephant Specification Language (ESL) \cite{Wilschut2024}, extracts dependencies from structured natural language, improving traceability and reducing ambiguity. ESL supports four of the five DSM-based analysis steps defined by Eppinger and Browning \cite{eppinger&browning2012}, but the critical first step—system decomposition—remains manual, time-intensive, and inconsistent.

Advances in natural language processing (NLP), particularly Large Language Models (LLMs) such as OpenAI’s GPT series, offer a potential breakthrough. These models extract dependencies and generate structured outputs from unstructured data sources \cite{bengio2021, bubeck2023}, producing possibilities in dependency mapping for MBSE workflows. Automating the initial DSM draft reduces reliance on expert elicitation while retaining human oversight, shifting engineers from full decomposition to targeted validation.

LLMs mitigate longstanding challenges in automated DSM construction by handling terminology variability \cite{bengio2021, bubeck2023}, extracting dependencies from unstructured sources \cite{eppinger&browning2012, Koh2024}, flagging uncertain dependencies through abstention responses (“I don’t know”) for expert validation \cite{Koh2024}, enhancing consistency in line with MBSE principles \cite{madni2018}. Although poor documentation quality remains an obstacle \cite{rubens2002}, contextual inference, explicit uncertainty marking, and pooling information across multiple documents help mitigate its impact \cite{Koh2024}. These advances rebalance the speed–quality trade-off identified in \cite{dong2002}, reducing weeks of preparation \cite{wilschut2018} to minutes \cite{Koh2024}, while enabling aggregation across sources for more robust analysis.

Despite these advances, research on LLM-based DSM generation remains in its early stages. To date, Auto-DSM \cite{Koh2024} is the only academic attempt to apply LLMs for automated system decomposition within the DSM domain, demonstrating the feasibility of this radical approach. However, its evaluation was case-specific, relying on a diesel engine handbook compared against an expert-derived DSM, where unequal information sources undermine reported accuracy. Reported results focused on completeness, with no validation of correctness, and only point metrics were provided. Without distributional analysis, stability and reproducibility across runs cannot be assessed—an essential requirement for dependable MBSE workflows. Moreover, the absence of controlled datasets prevents systematic variation of phrasing, complexity, and scope, making it impossible to isolate failure modes or establish benchmarks. Consequently, fundamental questions of accuracy, reproducibility, and reliability remain open, limiting trust and hindering the integration of LLM-based DSM tools into engineering practice. To address these gaps, this paper makes the following contributions:
\begin{enumerate}
    \item \textit{Black-box evaluation framework:} A systematic methodology to assess Auto-DSM by comparing LLM-generated DSMs against validated ground-truth DSMs under controlled and reproducible experimental conditions. 
    \item \textit{Extended evaluation metrics:} Introduction of new single-run and multi-run metrics that move beyond accuracy, capturing abstention behavior, incorrectness penalties, composite quality, and reliability across runs. 
    \item \textit{Systematic identification of limitations:} Application of the framework to both synthetic and real-world datasets provides the first reproducible evidence of Auto-DSM’s constraints, including reproducibility gaps and performance variability, thereby guiding future LLM-based DSM automation. 
\end{enumerate}
The evaluation framework and datasets are made publicly available at \url{https://github.com/nielspotters/Auto-DSM_Evaluation_framework}.

The remainder of this paper is organized as follows. Section~II presents the preliminaries. Section~III introduces the black-box evaluation methodology. Section~IV details the experimental setup. Section~V reports results. Section~VI discusses limitations and future work. Section~VII concludes the paper.

%% file: 2_Background_and_Problem_Definition/Background_and_Problem_Definition.tex
\section{Preliminaries}
\label{sec:Preliminaries}
This section reviews the key challenges in applying LLMs for DSM construction, presents preliminary experiments on Auto-DSM, and formulates the research problem.

\subsection{Hallucinations}
A key challenge in this context is hallucination—when the LLM generates confident but unsupported or incorrect dependencies. Such errors may arise from intrinsic factors (e.g., model bias, incomplete training data) or extrinsic ones (e.g., ambiguous or noisy input) \cite{Huang2025, ziwei2023}. In systems engineering tasks, hallucinations often manifest as fabricated or omitted DSM links. These outputs are typically expressed with high linguistic confidence, making them difficult to detect without reference to ground truth. Koh \cite{Koh2024} attempted to mitigate this by configuring the model with temperature = 0 and enabling abstention responses (“I don’t know”). However, empirical results show that this option is unreliable, as the model may assert a dependency in one direction while rejecting its symmetric counterpart—an inconsistency also observed in Koh’s example cases, where correctness never reaches 100\% across symmetric relations \cite{Koh2024}. 

Existing literature emphasizes that LLMs exhibit a systematic non-abstention preference: by default, models are more likely to produce an answer than to abstain, even when uncertain \cite{Wen2025, yadkori2024}. This tendency introduces risks in engineering applications, where an unsupported answer may appear as a valid system dependency. Several strategies have been proposed to counter this bias, including conformal abstention with theoretical guarantees \cite{yadkori2024}, strict prompting, verbal confidence thresholding, and Chain-of-Thought reasoning \cite{madhusudhan2024}. These approaches demonstrate that abstention behavior can be influenced through prompting or post-processing. Yet, no current Auto-DSM work explicitly evaluates abstention and confidence calibration under controlled test cases. As a result, claims about the reliability of “I don’t know” responses (e.g., via temperature tuning \cite{Koh2024}) remain unevaluated. Establishing a baseline assessment of abstention behavior is therefore essential, both to validate current prompt instructions and to determine whether advanced abstention strategies can improve the consistency and accuracy of LLM-generated DSMs. 

Koh reports only point performance metrics without any distributional analysis—leaving the stability and reproducibility of outputs unexamined. This omission is problematic in engineering contexts, where design decisions require repeatable and verifiable outputs. Similar concerns have been raised in adjacent fields: Colas et al. \cite{colas2018} demonstrated that, in reinforcement learning, statistically reliable conclusions only emerged after 15 to 50 independent runs, depending on effect size and variance. Likewise, Liem and Panichella \cite{Liem2020} argued that randomization without proper control undermines reproducibility in predictive software engineering. Without explicitly quantifying output variability, hallucinations may remain undetected and the robustness of LLM-generated DSMs cannot be assured.

\subsection{Current State Experiments}
While the Auto-DSM approach shows promise, the method remains immature and requires refinement before adoption in industrial workflows. Initial tests using the available no-code implementation revealed inconsistencies, as the code failed to reproduce the results reported in Koh’s study.

The first experiment assessed the sensitivity of Auto-DSM to terminology. Using the Heavy Duty Diesel Engines dataset \cite{Lakshminarayanan2020} from Koh’s example, three system parameter terms—“Diesel engine,” “Combustion engine,” and “Engine”—were compared, with dependency type set to “Mechanical.” The system identified only six elements across these definitions, whereas Koh reported eleven elements for an unspecified system parameter \autoref{tab:preexp}. Defining no dependency type resulted in ten elements for “Engine,” but the remaining definitions did not change. This indicates both terminology sensitivity and internal inconsistency, as the first pipeline prompt does not incorporate dependency type. These findings underscore the need to better understand LLM behavior in system decomposition before integration into automated workflows.

\begin{table}[t]
\centering
\caption{Comparison of system definitions and elements between Koh \cite{Koh2024} and empirical results}
\label{tab:preexp}
\begin{tabular}{lcc}
\toprule
\textbf{Source} & \textbf{Koh (2024) \cite{Koh2024}} & \textbf{Empirical results} \\ 
\midrule
Element 1 & Air-cleaner & Cylinder head \\
Element 2 & Compressor & Crankcase \\
Element 3 & Turbocharger & Connecting rod \\
Element 4 & Intake manifold & Pistons \\
Element 5 & Air cooler & Camshaft \\
Element 6 & Fuel injectors & Crankshaft \\
Element 7 & Cylinders & – \\
Element 8 & Exhaust gas turbine & – \\
Element 9 & ECU (Engine Control Unit) & – \\
Element 10 & Engine interface & – \\
Element 11 & Monitor & – \\
\bottomrule
\end{tabular}
\end{table}

A second analysis evaluated whether the workflow reduced hallucinations. The Auto-DSM approach attempted to enforce consistency by setting the model’s temperature to 0 and allowing “I don’t know” responses. However, this did not yield stable outputs. Across ten runs, the lowest reproducibility for a DSM entry was only 60\% (\autoref{tab:frequency}). While some variability reflected the model’s uncertainty, this was true for only two entries (highlighted in \autoref{tab:most-occurring} and \autoref{tab:idk-counts}). In many cases, the model alternated between “yes” and “no,” indicating hallucinations (see \autoref{sec: Appendix A}). Notably, the model was 100\% confident about the dependency from camshaft to crankcase, yet rejected the symmetric relation in 90\% of runs, again revealing inconsistencies.

\subsection{Problem Statement}
These preliminary tests highlight the need for deeper analysis of system decomposition in the proposed LLM pipeline and for assessing its potential in engineering practice. In particular, experiments with the Heavy Duty Engines dataset \cite{Lakshminarayanan2020} demonstrated that Auto-DSM is sensitive to terminology, exhibits inconsistent abstention behavior, and fails to reproduce results across repeated runs. Yet, due to the extensiveness of this dataset, such analysis is analytically intensive and difficult to interpret, as the reasoning of the LLM spans large volumes of information. 

Moreover, as in Koh’s evaluation, comparing Auto-DSM outputs with expert-derived DSMs is methodologically unsound, since the LLM was provided with a handbook while the expert DSM was produced independently. Koh implicitly assumes that both sources capture the same system knowledge, but this is neither stated nor verified. To properly assess Auto-DSM performance, a controlled dataset with an explicit ground-truth DSM is required, ensuring that both the model and the reference rely on the same knowledge base.

To address these limitations, this research adopts a controlled experimental design, constraining both data input and parameters to enable systematic evaluation of performance and limitations. While advances in DSM visualization and analysis have streamlined their use, the building process itself remains a bottleneck, constrained by manual effort and inconsistencies in data and communication. Existing model-based approaches, such as ESL, alleviate some of these challenges but do not automate the decomposition of system elements and dependencies. Similarly, while LLMs show potential in this direction, current methods lack the maturity and validation required for industrial adoption. 

This research therefore seeks to bridge this gap by designing and applying a black-box evaluation framework for LLM-based DSM system decomposition. The framework is tested on an existing Auto-DSM pipeline to provide a baseline assessment of current abstention and reliability, while establishing the foundations for future improvements. These objectives lead to the following research question: \\

\textbf{How can a black-box evaluation framework be designed to systematically assess LLM-based system decomposition for DSM generation, specifically in terms of identifying components and their dependencies?} \\

To address this research question the following sub-questions are formulated:
\begin{itemize}
    \item What evaluation criteria and metrics, grounded in comparison with validated ground-truth DSMs, can be used to assess the quality, coverage, and inaccuracies of LLM-based system decompositions, including the reliability of abstention behavior?
    \item How can controlled experimental setups—varying input phrasing, dataset scope, and system complexity—be designed to evaluate the accuracy, reproducibility, and hallucination behavior of current Auto-DSM methods, thereby establishing a baseline assessment of their performance?
\end{itemize}

%% file: 3_Methodology/Methodology.tex
\section{Methodology}
This section outlines the methodological framework designed to evaluate the ability of large language models (LLMs) to generate Design Structure Matrices (DSMs) from structured textual input. Building on prior work in design automation and language model evaluation, the approach is grounded in black-box testing principles, where only the input and output of the DSM generation pipeline are observable and used for assessment.

\subsection{Black-Box Evaluation of the Auto-DSM Pipeline}
The Auto-DSM pipeline is evaluated using a black-box testing approach, motivated by the closed nature of its implementation. The publicly released no-code version by Koh \cite{Koh2024} provides no access to internal logic, including prompt structure, vector storage, model embeddings or intermediate reasoning steps. As such, evaluation must be based solely on the observable relationship between structured input and DSM output.

Black-box testing is well suited for this context, as it focuses on validating functional behavior without reference to internal architecture or processing logic \cite{Beizer2003}, \cite{myers2011}. It is widely applied when system internals are inaccessible or abstracted, and is commonly used to detect incorrect functionalities, interface inconsistencies, and structural errors. Pressman \cite{Pressman2005} identifies five key categories of errors that black-box testing is intended to uncover: (1) incorrect or missing functions, (2) interface errors, (3) data structure or external database access issues, (4) behavioral or performance errors, and (5) initialization and termination faults. Among these, the first and fourth are particularly relevant for evaluating the Auto-DSM pipeline, where the core objective is to verify whether the system correctly infers component dependencies (i.e., functional correctness) and does so consistently across multiple executions (i.e., behavioral stability) from structured textual input.

In the proposed setup, Auto-DSM receives structured system descriptions as input and outputs a Design Structure Matrix (DSM). The evaluation framework compares the generated DSM (GEN-DSM) to a manually derived ground truth (GT-DSM), focusing on accuracy, consistency, and variability. To account for reproducibility, the system is executed N times per input, enabling measurement of output variability and supports quantification of stochastic behavior in the Auto-DSM pipeline. 

\subsection{Evaluation Framework}
The evaluation framework consists of two parts, first calculating the metrics for a single run based on the DSM characteristics and comparing the single run GEN-DSM with the GT-DSM. Secondly calculating the aggregated metrics based on $N$ runs. For both the single and aggregated comparison metrics visualization graphs are used and build into the evaluation framework to improve analysis utilizing different type of heatmaps.

\subsubsection{GT-DSM vs. GEN-DSM: Single-Run Evaluation}
\hspace{1cm}\vspace{0.1cm}\\
The single-run evaluation framework consists of four sequential stages: DSM alignment, metric computation, visualization, and structured output generation. Its purpose is to quantitatively assess the agreement between a generated DSM (GEN-DSM) and a manually curated ground truth DSM (GT-DSM), establishing a reproducible benchmark for model performance.

A key step is reconciling discrepancies in component labels between the GT and GEN matrices, which may result from formatting inconsistencies or semantic misinterpretations by the LLM. Initial alignment is performed via fuzzy string matching (token sort ratio) using the \texttt{RapidFuzz} library, with 100\% matches accepted automatically. For ambiguous cases, semantic similarity is computed using the \texttt{all-MiniLM-L6-v2} transformer-based encoder, pretrained on over one billion sentence pairs and widely applied in semantic retrieval \footnote{\url{https://huggingface.co/sentence-transformers/all-MiniLM-L6-v2}}. The highest semantic match from the GEN-DSM is proposed for user confirmation; if rejected, the user removes one of the conflicting components. The resulting matched set forms the basis for all subsequent computations, with evaluation restricted to components common to both matrices.

Each DSM entry is drawn from ${-1, 0, 1}$, representing absent, uncertain, and present dependencies, respectively. Three DSM-specific metrics—\textit{completeness}, \textit{correctness}, and \textit{Non Zero Fraction (NZF)}—provide macro-level structural assessment, following \cite{Koh2024, holtta2007}. Completeness is the fraction of entries classified as active or inactive, excluding uncertain values. Correctness is the proportion of active dependencies that are reciprocated, indicating structural symmetry. NZF, equivalent to coupling density, is the proportion of active dependencies relative to all possible off-diagonal entries, normalized for system size. These metrics are formally defined using Iverson bracket notation, where $[P]=1$ if predicate $P$ is true and $0$ otherwise:

\begin{equation}\label{eq:completeness}
\text{Completeness} = \frac{\sum_{i=1}^{N} \sum_{j=1,\, j \ne i}^{N} [\text{DSM}_{ij}\in\{1,-1\}\bigr]}{N(N - 1)}
\end{equation}

\begin{equation}\label{eq:correctness}
\text{Correctness} = \frac{\sum_{i=1}^{N} \sum_{j=i+1}^{N} [\text{DSM}_{ij} = 1 \land \text{DSM}_{ji} = 1]}{\sum_{i=1}^{N} \sum_{j=1,\, j \ne i}^{N} [\text{DSM}_{ij} = 1]}
\end{equation}

\begin{equation}\label{eq:nzf}
\text{NZF} = \frac{\sum_{i=1}^{N} \sum_{j=1,\, j \ne i}^{N} [\text{DSM}_{ij} = 1]}{N(N - 1)}
\end{equation}

Here, $N$ is the total number of system elements, and $i,j$ index the row and column positions within the $N\times N$ matrix. Diagonal entries ($i=j$) represent self-dependencies and are excluded.

While these metrics capture overall structural consistency, they do not measure prediction accuracy at the individual cell level. To address this, we incorporate \textit{Selective Accuracy}, a classification-based metric grounded in confusion matrix theory. This approach, common in supervised learning and black-box model auditing \cite{fawcett2006, pedregosa2011scikit}, treats each DSM cell as an independent classification into one of three states: presence $(1)$, absence $(-1)$, or uncertainty $(0)$. The inclusion of the uncertainty state introduces a rejection option, which is evaluated through \textit{Certainty Coverage}. In this context, Certainty Coverage is equivalent to the previously defined \textit{completeness} metric. Predictions are compared element-wise to the GT-DSM.

Let $\text{TP}$, $\text{FP}$, $\text{TN}$, $\text{FN}$, and $\text{IDK}$ denote true positives, false positives, true negatives, false negatives, and uncertain predictions, respectively, with $\text{Total}=N(N - 1)$. \textit{Selective Accuracy} is then defined as: 

\begin{equation}\label{eq:SA}
\text{Selective Accuracy (SA)} = \frac{\text{TP} + \text{TN}}{\text{Total} - \text{IDK}}
\end{equation}

This follows the paradigm of selective classification, or classification with a reject option, where accuracy is conditioned on the model choosing to respond \cite{elyaniv2010, geifman2019}. In this framework, Selective Accuracy measures correctness among committed predictions, whereas Certainty Coverage quantifies the proportion of the DSM for which the model abstains from prediction.

In combination, \textit{Selective Accuracy} and the three Product Architect DSM metrics—\textit{Completeness}, \textit{Correctness}, and \textit{NZF}—constitute the single-run evaluation framework. This framework enables transparent, reproducible, and interpretable comparison between the GEN-DSM and GT-DSM. It evaluates a single deterministic output and does not capture stochastic variability across multiple runs, which is addressed in the multi-run evaluation (Section~\ref{subsubsec:Multirun}).

For interpretability, the framework generates three visual diagnostics: (i) a matched DSM heatmap showing correct predictions, (ii) a mismatch DSM heatmap highlighting errors, and (iii) an uncertainty DSM heatmap marking the rejected response entries. All are exported in rasterized (PNG) and tabular (Excel) formats for both qualitative review and quantitative analysis.\\

\subsubsection{Multi-Run Evaluation and Composite Quality Aggregation}\label{subsubsec:Multirun}
\hspace{1cm}\vspace{0.1cm}\\
While the single-run evaluation framework offers insights into the correctness of a single LLM-generated DSM instance, it does not capture the reproducibility or behavioral boundaries of the model. A core objective of this research is to assess not only the classification performance, but also the consistency and variability of the DSM generation pipeline. In line with best practices in empirical NLP evaluation \cite{dror2018,bouthillier2019}, we systematically analyze the outputs of repeated pipeline executions under identical input conditions to quantify variability and robustness. Despite setting the decoding temperature to zero, non-determinism in LLM inference can arise from parallel computation artifacts, floating-point instability, or token alignment ambiguities \cite{belinkov2017}. Therefore, we execute the Auto-DSM pipeline $N$ times for a fixed input and evaluate the resulting $N$ DSMs collectively. Before any performance metrics are computed, the number of repeated Auto-DSM executions (denoted as $N$) must be determined in a statistically grounded manner. Rather than reporting agreement with the Ground Truth DSM (GT-DSM) based on a single instance---thus risking results influenced by chance---it is essential to estimate a confidence interval for the aggregated agreement across $N$ independent runs. This aligns the evaluation with established principles of statistical reliability. To generate an aggregated agreement, size differences between generated DSMs have to be adressed due to inconsistent component identification.

\paragraph{Label Harmonization and Alignment} Each generated DSM may contain a distinct set of identified system components. To compare across runs, all DSMs are realigned to a common label set $\mathcal{L} = \bigcup_{i=1}^{N} L_i$, where $L_i$ is the label set extracted in run $i$. This union ensures structural comparability without discarding model output. Each matrix is reindexed using $\mathcal{L}$, with missing rows/columns filled using \texttt{NaN} to indicate absence of the dependency, rather than a semantic zero (which carries the specific meaning of uncertainty in the Auto-DSM context). Let $\mathcal{V}_{rc} = \{ D^{(1)}_{rc}, D^{(2)}_{rc}, \ldots, D^{(N)}_{rc} \}$ denote the non-\texttt{NaN} values at cell $(r,c)$ across $N$ aligned DSMs.

\paragraph{Agreement Criterion and Sample-Size Policy}
To evaluate the reproducibility and consistency of model predictions across multiple executions of the Auto-DSM pipeline, we adopt a two-pronged agreement analysis based on (1) the raw agreement rate and (2) Fleiss’ $\kappa$ statistic.

The \textit{raw agreement rate} measures the proportion of identical predictions across multiple runs for each cell in the DSM. For every off-diagonal cell $(i,j)$, we collect a prediction vector over $N$ runs as $v_{ij} = [x_1, x_2, \ldots, x_N]$, where $x_k \in \{-1, 0, 1\}$ denotes confidently absent, uncertain, or confidently present dependencies, respectively. The agreement rate for a given cell is defined as the proportion of predictions belonging to the majority class. The DSM-wide agreement is then summarized by the \textit{mean} and \textit{standard deviation} of all valid cell-level agreement scores.

To account for agreement by chance and better capture categorical reliability, we compute \textit{Fleiss’ $\kappa$} on a per-cell basis \cite{fleiss1979}. Each off-diagonal cell $(i,j)$ is treated as having $N$ independent categorical labels. Fleiss' $\kappa$ for each cell is given by:

\begin{equation}
    \kappa_{ij} = \frac{P_{ij} - P_e}{1 - P_e}, \quad 
    P_{ij} = \frac{\sum_c n_c (n_c - 1)}{N(N - 1)}, \quad
    P_e = \sum_c p_c^2,
\end{equation}

where $n_c$ is the count of label $c \in \{-1, 0, 1\}$ in the cell $(i,j)$, and $p_c$ is the marginal probability of class $c$ across all predictions and cells. Only off-diagonal cells with at least five valid predictions are included to ensure reliable estimation. The overall agreement across the DSM is then computed as the average over all such valid cells:

\begin{equation}
    \bar{\kappa} = \frac{1}{|\mathcal{C}|} \sum_{(i,j) \in \mathcal{C}} \kappa_{ij},
\end{equation}

where $\mathcal{C}$ denotes the set of all off-diagonal cells with sufficient observations.

To establish engineering-grade reproducibility, we adopt a conservative threshold of $\bar{\kappa} \geq 0.60$, corresponding to ``substantial agreement'' on the Landis–Koch scale \cite{landis1977}. This threshold is motivated by empirical findings in DSM literature, where inter-rater disagreement among human experts is nontrivial \cite{Tilstra2012, Schmitz2011}.

An initial sample size of $N = 30$ pipeline runs is used. If the resulting average agreement $\bar{\kappa} < 0.60$, we incrementally increase the sample size by 10 runs until either the threshold is met or a maximum of $N = 60$ is reached. This adaptive sampling approach is guided by empirical findings from stochastic system evaluations, which show that reproducible agreement signals typically emerge within 15–50 independent executions \cite{colas2018}. 

\paragraph{Match and Mismatch Frequencies} For each cell, we compute the match frequency $M_{rc}$ and mismatch frequency $\bar{M}_{rc}$:
\begin{align}
    M_{rc} &= \frac{1}{|\mathcal{V}_{rc}|} \sum_{x \in \mathcal{V}_{rc}} [x = \text{GT}_{rc}], \\
    \bar{M}_{rc} &= 1 - M_{rc}.
\end{align}
These are derived from the matched and mismatched heatmaps, which encode the cell-wise alignment of the generated DSM against a reference ground-truth DSM based on the results from the single-run evaluation.

\paragraph{Uncertainty Frequency} The frequency of uncertain responses (i.e., model output $0$) is given by:
\begin{equation}
    U_{rc} = \frac{1}{|\mathcal{V}_{rc}|} \sum_{x \in \mathcal{V}_{rc}} [x = 0],
\end{equation}
which measures the likelihood that the model abstains from a confident prediction for dependency $(r,c)$.

\paragraph{Per-Cell Aggregation and Entropy} In the context of information theory, the entropy per cell for a variable taking values in \(\{-1, 0, 1\}\) is calculated based on the discrete probability mass function over these three outcomes. Let \(p_{-1}\), \(p_0\), and \(p_1\) denote the probabilities of observing the values \(-1\), \(0\), and \(1\), respectively, such that \(p_{-1} + p_0 + p_1 = 1\).  The variability of predictions per cell is quantified using Shannon entropy:
\begin{equation}
    H_{rc} = -\sum_{v \in \{-1, 0, 1\}} p_v \cdot\log_2 p_v,
\end{equation}
where $p_v = \frac{1}{|\mathcal{V}_{rc}|} \sum_{x \in \mathcal{V}_{rc}} [x = v]$ and $[P] = 1$ if $P$ is true, 0 otherwise. As in standard entropy definitions, terms with zero probability do not contribute to the sum due to the convention that \(0 \log 0 := 0\), justified by the limiting behavior \(\lim_{p \to 0^+} p \log p = 0\). This yields values in the range $[0, \log_2 3]$, since the maximum entropy occurs when $p_{-1} = p_0 = p_1 = \frac{1}{3}$.

To make the stability term in the Composite Quality Score $Q$ directly comparable across systems, we normalize the entropy to a $[0,1]$ range:
\begin{equation}
    H^{\text{norm}}_{rc} = \frac{H_{rc}}{\log_2 3}.
\end{equation}
Where, $H^{\text{norm}}_{rc} = 0$ indicates perfect stability (all runs gave the same value for the cell) and $H^{\text{norm}}_{rc} = 1$ indicates maximum instability (the three possible values are equally likely).

This ensures mathematical consistency and avoids undefined expressions. The entropy $H_{rc}$ thus captures the uncertainty of the model per matrix entry, with higher values indicating greater prediction instability. For matrix cells with high entropy but low participation (i.e., high \texttt{NaN} frequency), interpretability is enhanced by computing the fill ratio $f_{rc} = |\mathcal{V}_{rc}| / N$.

\paragraph{Selective accuracy per cell} The selective accuracy per call is given by:
\begin{equation}
    SA_{rc} = \frac{M_{rc}}{1-U_{rc}}
\end{equation}
which defines the correctness of the confident responses of the model. Since $\bar{M_{rc}}$ contains both the confident as uncertainty responses, $U_{rc}$ has to be subtracted getting the selective accuracy (i.e. the accuracy without uncertainty, GEN = $0$ responses).

\paragraph{Confidence Penalty}
While these metrics capture aspects of correctness and robustness, they assume uniform cost for different types of errors. This assumption is problematic in engineering design contexts, where the impact of errors varies significantly. Specifically, false positives may lead to unnecessary design coupling and overengineering, while false negatives can result in undetected dependencies that compromise system integration \cite{browning2002}. Uncertain predictions ($\text{IDK}$) offer limited engineering value, as they leave ambiguity in critical design decisions. Uncertain predictions are preferable to incorrect ones, as they signal ambiguity and allow for focused investigation by engineers since they are clearly represented in the responses. In contrast, confidently incorrect outputs are harder to detect in the absence of a ground truth DSM—a gap the Auto-DSM pipeline aims to address by enabling DSM generation from unstructured textual descriptions. To address these concerns, we adopt a cost-sensitive error model inspired by \cite{elkan2001foundations}, in which distinct penalties are assigned to different prediction types:

\begin{equation}
    P_{rc} = \frac{1}{|\mathcal{V}_{rc}|} \left( \sum_{x \in \mathcal{V}_{rc}} w_{inc} \cdot[x \neq 0 \land x \neq \text{GT}_{rc}] + w_{idk} \cdot [x = 0] \right).
\end{equation}
Where $w_{inc} =1$ and $w_{idk} = 0.5$, this formulation penalizes false predictions most heavily, reflecting their disruptive impact on system architecture. Abstentions incur a smaller penalty, consistent with selective classification frameworks \cite{chow2003, elyaniv2010}, where cautious behavior is preferable to confidently incorrect outputs. Weight factors can be adjusted dependent on domain or case.

\paragraph{Composite Quality Score (Q)} To synthesize these indicators into a single quality measure, we define a Composite Quality Score (Q) that aggregates accuracy, stability, and penalization into a single metric, we define:
\begin{equation}
    Q_{rc} = w_{\text{acc}} \cdot SA_{rc} + w_{\text{stab}} \cdot (1 - H_{rc}) - w_{\text{pen}} \cdot P_{rc},
\end{equation}
The weights \( w_{\text{acc}} = 0.5 \), \( w_{\text{stab}} = 0.3 \), and \( w_{\text{pen}} = 0.2 \) reflect engineering priorities where correctness is paramount, stability is desirable, and error severity must be penalized but not overly dominant. This weighted composite score serves as a principled starting point, consistent with established practices in multi-objective evaluation \cite{hand2009measuring}, and addresses known limitations of single-metric benchmarks in ML \cite{bouthillier2021accounting}.

\paragraph{Normalized Composite Quality Score} Since the Q is a linear combination with a subtractive penalty term, its raw score is not inherently bounded between 0 and 1. To facilitate interpretability and cross-system comparison, we normalize the score:
\begin{equation}
    Q_{\text{norm},rc} = \frac{Q_{rc} - Q_{\min}}{Q_{\max} - Q_{\min}},
\end{equation}
where $SA_{rc} \in [0,1]$, $H_{rc} \in [0,1]$, and $P_{rc} \in [0,1]$. The theoretical bounds are given by:
\begin{align}
    Q_{\max} &= w_{\text{acc}} + w_{\text{stab}}, \\
    Q_{\min} &= - w_{\text{pen}}.
\end{align}
This normalization ensures that the best-performing model achieves $Q_{\text{norm}} = 1$ (i.e., perfect accuracy, full stability, no penalty), while the worst case yields $Q_{\text{norm}} = 0$ (i.e., total failure across all metrics). Such linear normalization is commonly applied in composite metric design to support inter-model comparability, especially when integrating partially competing objectives \cite{chinchor1993, fawcett2006}. The use of $Q_{\text{norm}}$ provides a dimensionless and interpretable index of DSM generation quality, facilitating fair benchmarking of LLM performance across datasets.

\paragraph{Heatmap Visualization and Export} To enable rapid interpretation of the evaluation results, aggregated metric matrices are visualized as heatmaps. Each heatmap represents pairwise component relationships, allowing both global patterns and localized anomalies to be identified at a glance. Metrics such as match frequency, mismatch frequency, uncertainty frequency, normalized entropy, performance quality, average DSM values, fill ratio, and raw agreement are each generated, and exported in tabular (Excel) and visual (annotated heatmap) formats with indexed component labels. These visual diagnostics facilitate both macro-level system interpretation and component-level inspection of  LLM prediction behavior.

This multi-run framework enables empirical quantification of variability in model behavior during LLM-based DSM generation. By integrating structural, classification, and entropy-based metrics, it identifies regions of epistemic uncertainty and unstable dependency predictions—insights that are critical in engineering contexts where overlooked interactions may compromise system integrity. Beyond correctness, the framework captures consistency across executions, supporting rigorous validation of the Auto-DSM pipeline. It also lays the groundwork for iterative improvements across multiple fronts: prompt engineering, to refine input clarity and task interpretation; decoding control, to reduce stochasticity and stabilize output; ensemble-based generation, to aggregate multiple outputs for enhanced robustness; and data architecture, to ensure consistent formatting, labeling, and preprocessing of structured input across evaluations. Crucially, this approach avoids the common methodological pitfall of treating a single model output as representative—an assumption particularly hazardous in high-stakes design environments where system dependencies must be interpreted reliably.

%% file: 4_Experimental_Setup/Experimental_Setup.tex
\section{Experimental Setup}
The experimental setup for black-box evaluation of the Auto-DSM comprises three parts: (1) input variables, defining variability in parameter and dataset inputs; (2) executed experiments, which instantiate test cases to quantify how specific inputs affect LLM behavior; and (3) ground truth and technical-document dataset design, producing controlled, symmetric, and sound datasets for a given system.

\subsection{Variable Parameters}
The pipeline supports variability in both input parameters and dataset types. To design the experiments, this variability is first defined and structured. 

\subsubsection{System input definition}
The Auto-DSM pipeline uses two parameters to define the target system and the dependency type. These parameters are injected into the prompts for component and dependency identification, enabling a scoped analysis for a specified system decomposition. Because the parameters alter the prompt and thus the task posed to the LLM, different settings may yield different responses; their influence therefore warrants analysis. The parameters are open-ended to allow broad customization. In this study, we assume the engineer targets a specific system decomposition; at minimum, the system must be defined. Both systems and subsystems can be specified, enabling hierarchical DSM analysis. The dependency type may be omitted to generate a DSM from general interactions, or specified to analyze particular dependency categories; omission still permits topological analysis of the system via unspecified interactions.

\subsubsection{Dataset variation}
Whereas the system definition is straightforward, the dataset can vary along multiple dimensions. Technical documentation underlying the system decomposition may differ in phrasing of interactions as well as in the complexity of the decomposition itself. Both aspects are investigated to assess the effects of linguistic variation and decomposition complexity on accuracy and scalability.

\paragraph{Interaction Definition}\label{ID}
Component-to-component DSM dependencies can be classified into four possible definition types:
\begin{enumerate}
    \item Direct link: A is linked to B
    \item Indirect link: A is linked to C via B
    \item No link: C is not linked to D
    \item Undefined link: Link between A or B with D (based on combining previous three types).
\end{enumerate}
These definitions evaluate how the pipeline handles different phrasings of stated dependencies and help identify semantic reasoning issues arising from interaction formulation.

Beyond distinguishing interaction types, ambiguity and inaccuracies must also be considered. Traditional document-centric approaches have been shown to introduce inconsistencies that can lead to errors \cite{Dambietz2021}; the experiments therefore also evaluate robustness to such noise.

\begin{figure}[H]
    \centering
    \includegraphics[width=0.7\linewidth]{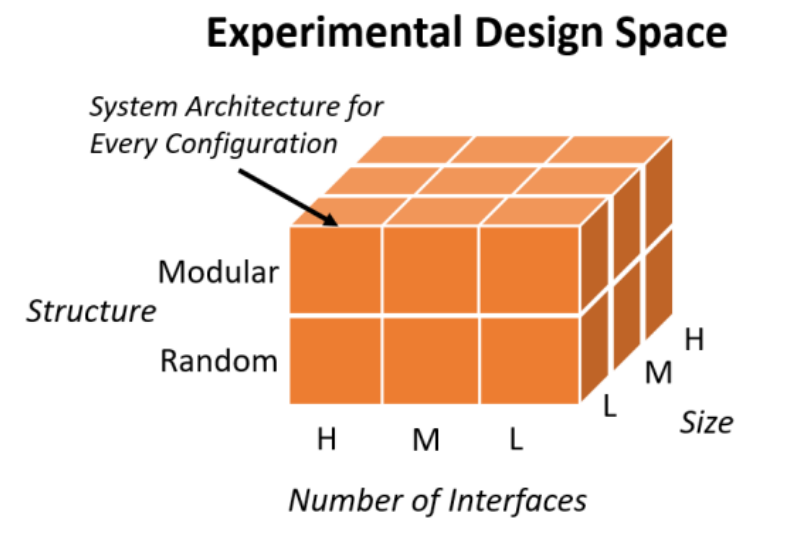}
    \caption{Experimental design space by Hennig et al.  \cite{hennig2021complexity}}
    \label{fig:EDSHennig}
\end{figure}

\paragraph{System complexity:}
In addition to interaction types, this study examines how variation in system decomposition affects Auto-DSM performance. Specifically, we evaluate how increasing data complexity influences the LLM’s ability to generate accurate DSMs. Following established benchmarks in the DSM literature, system complexity is assessed along three dimensions: system size (number of components), interconnectivity (number and density of interfaces), and architectural structure (e.g., degree of modularity or hierarchy), as visualized by Hennig et al. in the experimental design space in \autoref{fig:EDSHennig} \cite{hennig2021complexity}.

By systematically varying these decomposition characteristics together with the interaction definitions and system-input parameters, the study assesses the robustness, accuracy, and scalability of Auto-DSM under increasingly complex and noisy or mismatched input conditions. 



\subsection{Experiments}
Given the variability options, the experiments are designed to incorporate these variations and test pipeline behavior under each case. Specifically, we assess how selected input parameters and dataset properties influence Auto-DSM, with attention to inaccuracies, hallucinations, and known limitations. For every experiment, black-box testing is performed: a dataset and GT-DSM are generated, validated for correctness, refined if required for the specific case, input parameters are set, Auto-DSM is executed $N-$times, results are evaluated, and outcomes are integrated into the evaluation framework. A visual representation is illustrated in \autoref{fig:dataflow_experiment}. 

The experiments are grouped into three categories: 1) Interaction Definition, 2) System Parameters vs. Dataset, and 3) Complexity levels within the system decomposition. These categories isolate the effects of natural-language descriptions, inaccuracies in system and dataset design, and decomposition complexity on Auto-DSM performance. 

\subsubsection{Interaction Definition}
These experiments examine whether alternative phrasings of interactions affect pipeline responses. Datasets are based on a synthetic system in which interaction-type distinctions are omitted to isolate phrasing effects, resulting in the system parameters {Abstract system} with no defined dependency type. Three experiments are conducted:

\paragraph{Phrasing of interaction definitions}
The first experiment evaluates how the four interaction types—\textit{Direct, Indirect, No link, Undefined}—affect pipeline accuracy. An abstract system containing all four definitions enables a structured comparison. Performance for each interaction type is quantified using the evaluation framework.

\paragraph{Symmetry within interaction definitions}
The second experiment extends the first by mirroring all interactions so that every relation is bidirectional (i.e., \textit{Component A is linked to B; Component B is linked to A}). The goal is to assess whether bidirectional definitions influence DSM symmetry.

\paragraph{Inaccuracies within interaction definitions}
The third experiment measures the impact of inaccuracies by refining the abstract-system dataset with duplicated statements for direct interactions—either reinforcing the same relation (redundancy) or contradicting it (ambiguity). The evaluation framework is used, and performance is compared against the Uni-directional dataset to quantify the effect of inaccuracies. To assess redundancy, a component pair with high uncertainty is selected, whereas a component pair with high performance is chosen to evaluate ambiguity.

Based on these results, the most reliable interaction definitions are selected for subsequent dataset design. This choice focuses later experiments on variables of interest while maintaining higher accuracy and stability; lower-performing definitions may introduce excessive variance and confound analysis near the limits of Auto-DSM.

\subsubsection{System Parameters vs. Dataset}
The Auto-DSM pipeline accepts two input parameters (target system and dependency type) alongside any defined system decomposition. Because datasets and parameters are independently specified, they may match or mismatch. Both conditions are tested to quantify the influence of parameter–dataset alignment. Given the earlier assumption that engineers define the system, the cases in \autoref{tab:system_interactiontype} are considered.

\begin{table}[htbp]
\centering
\small
\begin{tabular}{|c|c|c|c|c|}
\hline
\textbf{} & \multicolumn{2}{c|}{\textbf{System}} & \multicolumn{2}{c|}{\textbf{Interaction type}} \\
\cline{2-5}
\textbf{} & \textbf{Parameter} & \textbf{Dataset} & \textbf{Parameter} & \textbf{Dataset} \\
\hline
1 & Defined & Included & Defined     & Included \\
2 & Defined & Included & Defined     & Excluded \\
3 & Defined & Included & Un-defined & Included \\
4 & Defined & Included & Un-defined & Excluded \\
5 & Defined & Excluded & Defined     & Included \\
6 & Defined & Excluded & Defined     & Excluded \\
7 & Defined & Excluded & Un-defined & Included \\
8 & Defined & Excluded & Un-defined & Excluded \\
\hline
\end{tabular}
\caption{Comparison of system and interaction type inclusion across parameters and datasets}
\label{tab:system_interactiontype}
\end{table}

In \autoref{tab:system_interactiontype}, \textit{Included} indicates that the defined parameter appears in the dataset; \textit{Excluded} indicates it does not. Experiments 4 and 8 warrant clarification: although seemingly counterintuitive, these use datasets for the defined (4) or another system (8) that contain only components and no interactions. These cases explore model responses when inputs are mismatched. To minimize interference from decomposition complexity, a low-complexity system is used. As before, GEN-DSM and GT-DSM are compared using the evaluation framework.

\subsubsection{Complexity within System Decomposition}
Complexity experiments are defined with the experimental design space in mind. Three cases are considered: a) increasing complexity within a subsystem, b) introducing hierarchy (two subsystems), and c) adding noise by including other systems alongside the defined system parameter.

\paragraph{Complexity within Subsystems}
Two subsystems with different complexity levels are selected based on the experimental design space. The highest- and lowest-complexity subsystems are characterized by size, inter-connectivity (computed via NZF, \autoref{eq:nzf}), and the modularity/integral structure of the GT-DSMs. The evaluation framework is then used to quantify the effect of subsystem complexity.

\paragraph{Complexity within hierarchy}
Because a system decomposition often spans multiple subsystems and components, hierarchical effects must also be evaluated. This experiment assesses not only intra-subsystem interactions but also how accurately the pipeline identifies components across both subsystems and the inter-subsystem interactions. Two subsystems (those used above) are provided while the system is defined as the pipeline input parameter, enabling comparison against the GT-DSM and assessment of whether hierarchy alters system-level performance relative to subsystem-level analysis. 

\paragraph{Complexity by Noise}
This experiment evaluates the extent to which irrelevant information—relative to the defined input variables—affects the generated DSM. Using the same dataset as the hierarchy experiment, the input variable is set to both the low- and high-complexity subsystems to measure how dataset noise influences pipeline performance.

\subsection{Dataset Design}\label{Dataset Design}
A Ground Truth (GT) dataset is required to evaluate pipeline responses with the evaluation framework. Using controlled datasets—where variables can be systematically manipulated—enables rigorous assessment of Auto-DSM, and repeated runs quantify deviations from the expected response.

Accordingly, system decompositions are specified in plain text, from which both the GT-DSM and experimental datasets are derived. To focus on the pipeline’s natural-language processing, only textual technical documents are used; images, audio, and model-based inputs are excluded. To keep the analysis compact and manually verifiable, two GT datasets are employed: \textit{1) a fictive abstract system} and \textit{2) a Refrigerator}.

\subsubsection{Fictive Abstract System Decomposition Dataset}\label{sub:fictive}
\begin{figure}[H]
    \centering
    \includegraphics[width=\linewidth]{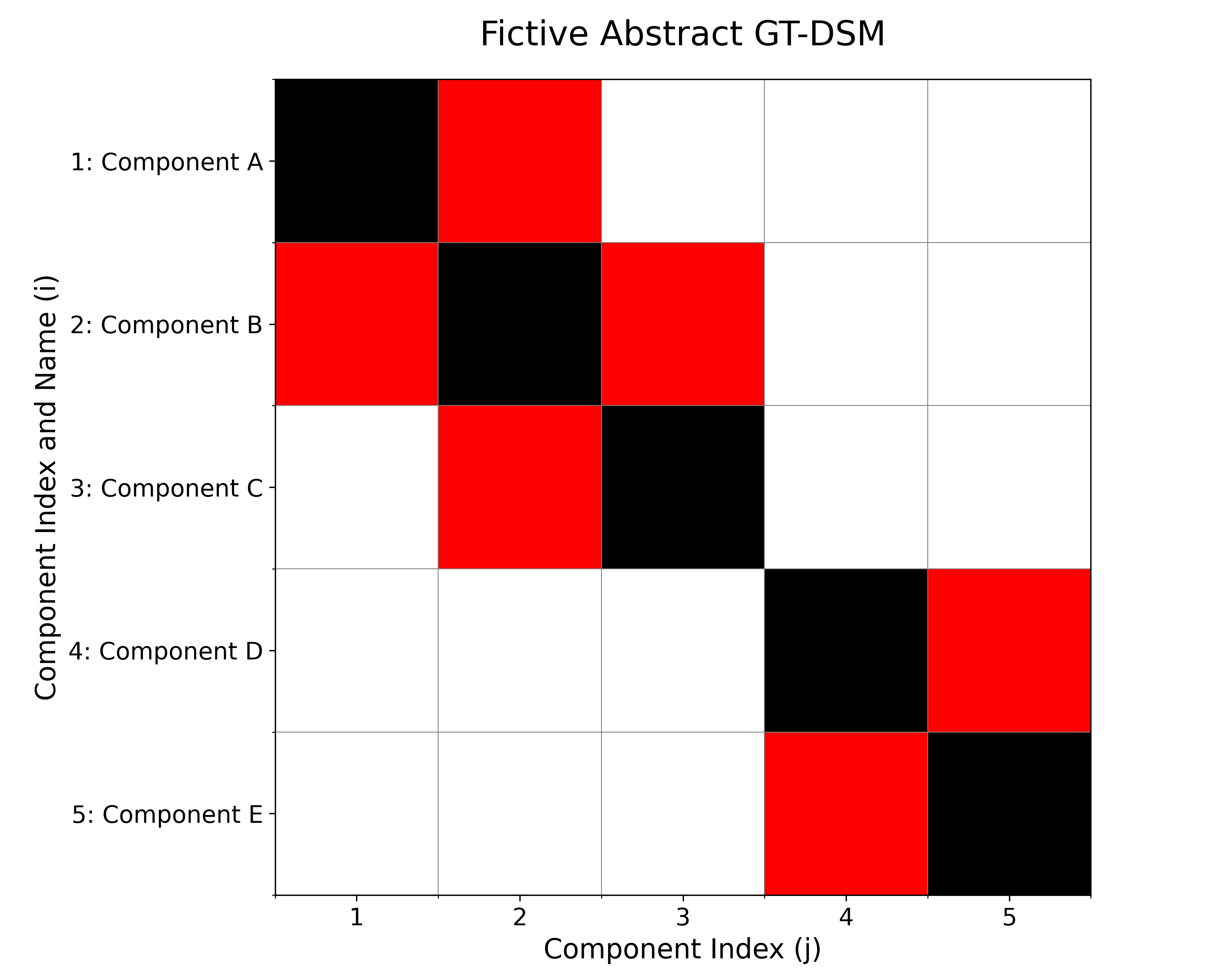}
    \caption{Ground Truth DSM of Fictive Abstract System}
    \label{fig:fictive GT-dsm}
\end{figure}

The fictive abstract system comprises 5 components, minimally constructed to cover the four interaction definitions described earlier. The decomposition is:
\begin{quote}\label{q:fictive}
    \textit{"An abstract system is composed of components A, B, C, D, and E. The components interact to ensure system functionality, as detailed below:
    \\
    Component A is linked to Component B. Component B is linked to Component C.
    Component C is not linked to Component D. Component D is linked to component E."}    
\end{quote}
Because the focus is phrasing, no dependency types are included—only components and interactions. The corresponding fictive abstract GT-DSM was manually constructed, resulting in \autoref{fig:fictive GT-dsm}.

This dataset is also used for the Bi-directional dataset, incorporating symmetry by stating every interaction twice in different sequence, i.e.: 
\begin{quote}\label{q:Bi-direc}
    \textit{"... Component A is linked to Component B. Component B is linked to Component A. ..."}    
\end{quote}

With minor refinements, the same dataset supports the duplication and contradiction experiments as well. The text is modified as follows:
\begin{itemize}
    \item 
    \textbf{Duplication:}
    \textit{"... Component D is linked to component E. Component D is linked to component E. ..."}    
    \item  \textbf{Contra-dictionary:}
    \textit{"... Component D is linked to component E. Component D is not linked to component E. ..."}    
\end{itemize}

Lastly, the Bi-directional dataset is refined to incorporate different dependencies for direct defined dependency pairs. This dataset is used for the Input Parameter vs Dataset experiments and is defined as: 
\begin{quote}
   \textit{"... Component A is \{dep-type\} linked to Component B. ..."}
\end{quote}
These interaction types are incorporated to investigate the accuracy of predicting specific interaction types as to left unspecified. The abstract system isolates the experiment to minimize effects of pretrained knowledge while employing two interaction types from the HDDSM framework described by Tilstra et al. \cite{Tilstra2012}.

\subsubsection{Refrigerator Decomposition Dataset}\label{sub:ref}
To evaluate LLM-based system decomposition and DSM generation, a system with analytical control, input validation, and a clearly bounded scope is required. While the abstract system is controllable and well-bounded, it lacks scalability, which is necessary for the input-parameter–vs–dataset and complexity experiments. A real system is therefore needed to assess decomposition performance on real-world data. Koh \cite{Koh2024} considered two candidates: a domestic refrigerator and a diesel engine. We select the domestic refrigerator as the primary ground truth because it has compact scale, well-defined subsystem boundaries, and manageable complexity. By contrast, Koh’s diesel engine represents a subsystem within a larger vehicle platform and lacks a clear system boundary, complicating component/interface selection, hierarchy definition, sizing relationships, and interconnectivity with surrounding subsystems. The refrigerator can be decomposed comprehensively—from product level to key components and interactions—without exceeding the intended scope. 

\paragraph{Refrigerator Ground Truth Dataset}
To generate the ground truth from the main components, we use access to ChatGPT (powered by OpenAI's Language Model, GPT-4o, \href{http://openai.com}{http://openai.com}) to efficiently propose generally accepted refrigerator decompositions based on its broad knowledge. 

\begin{figure}[t]
    \centering
    \includegraphics[width=\linewidth]{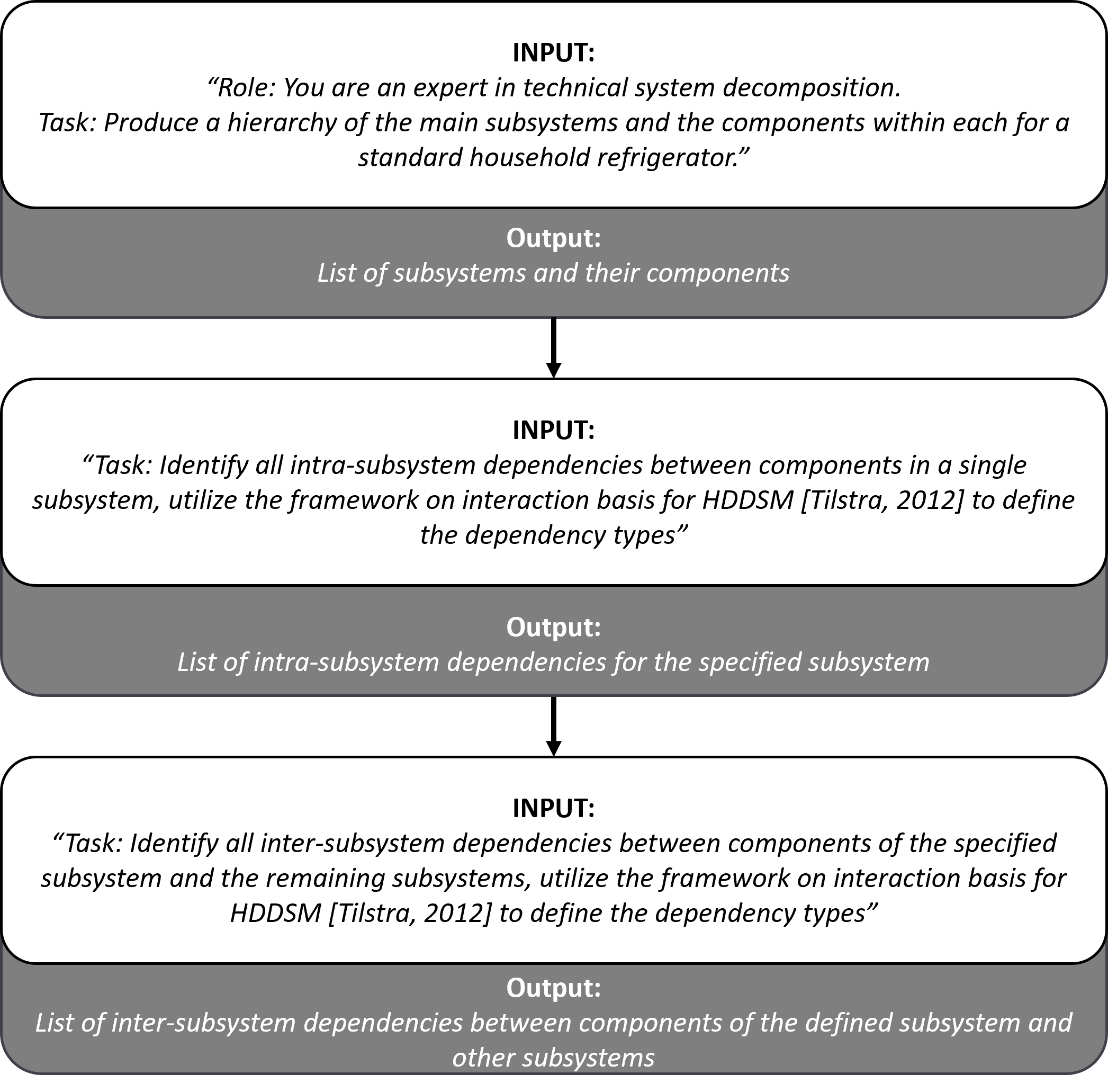}
    \caption{Ground Truth Dataset Generating Chain of Thought}
    \label{fig: GT-Pipeline}
\end{figure}

The ground truth decomposition was created using a structured prompt sequence, consisting of three sequential prompts where the prompts first extracts major subsystems, then main components per subsystem, and finally interactions. The interactions cover both intra- (within a subsystem) and inter-subsystem connectivity. \autoref{fig: GT-Pipeline} visualizes the Chain of Thought for the GT-decomposition, the actual used prompts can be found  in \autoref{lst:prompt1}, \autoref{lst:prompt2}, \autoref{lst:prompt3} of Appendix \ref{sec: Appendix A}.

All interactions are classified according to the High-Dimensional Design Structure Matrix (HDDSM) framework by Tilstra \cite{Tilstra2012}, widely adopted in engineering design research. This framework partitions interactions into five categories—\textit{Information, Material, Energy, Spatial, and Movement}—with finer subtypes (e.g., 1.2 Control [CI], 3.5 Electrical [EE]). The same framework is used in recent model-based systems engineering for DSM generation, such as the ESL framework by Wilschut et al. \cite{Wilschut2024}.

To ensure accuracy, reproducibility, and consistency across executions, the prompt structure incorporates the following strategies:

\begin{itemize}
    \item \textbf{Task Instruction Prompting:} 
    Each prompt clearly defines the model’s role, task, and scope. For instance, Prompt 1 begins with: \textit{``You are a systems engineer specialized in technical system decomposition and dependency analysis.''} 
    This technique helps align model outputs with domain expectations~\cite{White2023,Ekin2023}.

    \item \textbf{Constrained Generation:}
    Prompts include specific limitations such as: \textit{``Do not invent components or interaction types.''} 
    This restricts the LLM to predefined, verifiable terms—particularly in Prompt 2 and 3, which limit output to HDDSM interactions—thus reducing hallucination and improving consistency~\cite{White2023}.

    \item \textbf{Chain-of-Thought Structuring:}
    The decomposition task is split into sequential prompts: structure (Prompt 1), intra-subsystem interactions (Prompt 2), and inter-subsystem dependencies (Prompt 3). 
    This step-wise breakdown simulates reasoning chains, improving output reliability~\cite{Wei2023}.

    \item \textbf{Format-forcing Templates:}
    All prompts include strict output formats, such as: 
    \texttt{[Subsystem; From (Component); To (Component); Interaction Type(s); ...; Target Subsystem]}. 
    This reduces formatting ambiguity and enables direct parsing into DSMs~\cite{Ekin2023}.

    \item \textbf{In-context learning:}
    In-context learning relies on  examples closely related to the task, which the model uses to generalize to new tasks. I.e. \texttt{\# Example Template:
Subsystem 1; ...; Subsystem 1} \cite{brown2020language}.
    
\end{itemize}
\begin{figure*}[t]
  \centering
  \begin{minipage}{0.49\textwidth}
    \centering
    \includegraphics[width=\linewidth]{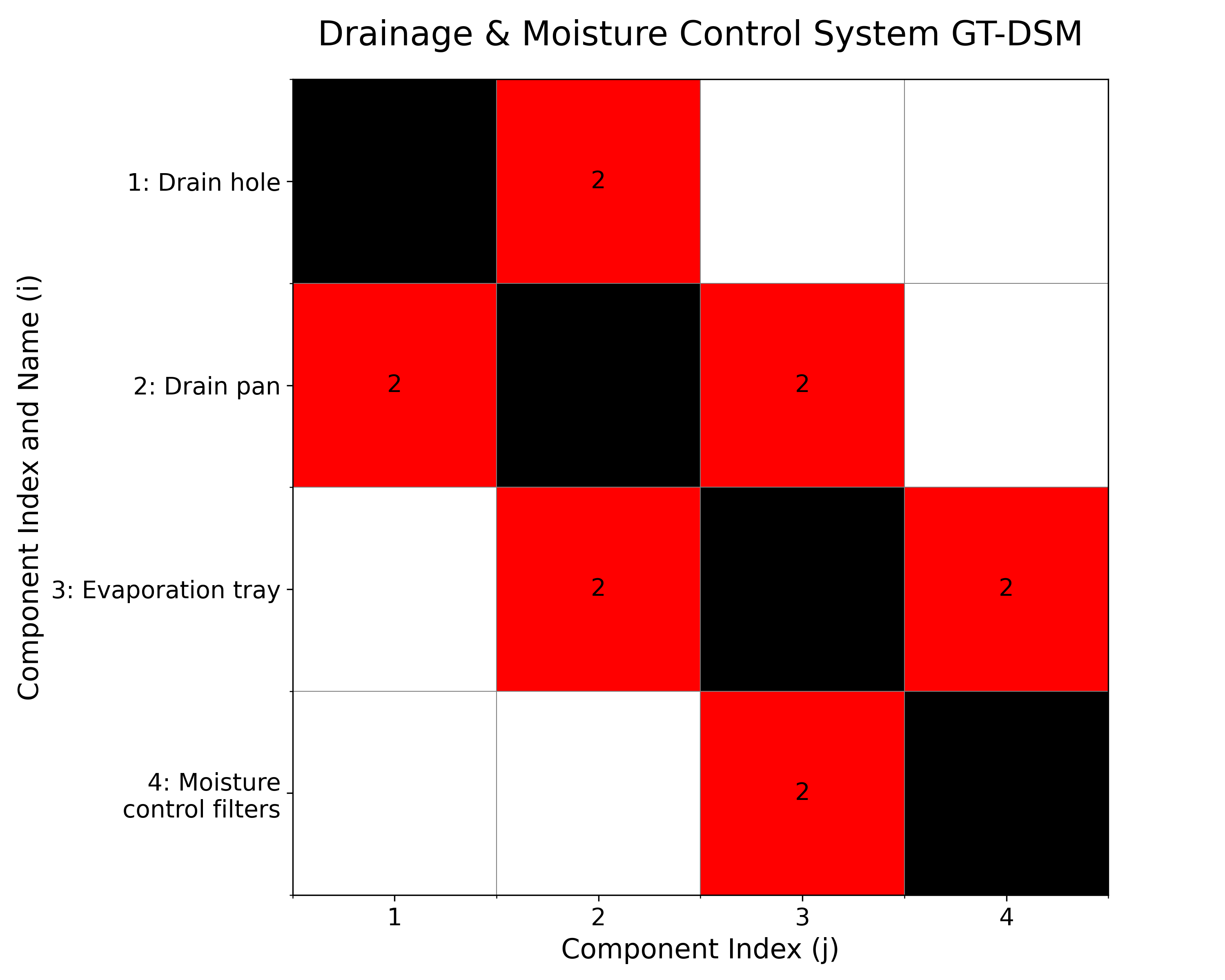}
    \caption{GT‑DSM of the Drainage \& Moisture Control System}
    \label{fig:GT-DSM:D&M}
  \end{minipage}\hfill
  \begin{minipage}{0.49\textwidth}
    \centering
    \includegraphics[width=\linewidth]{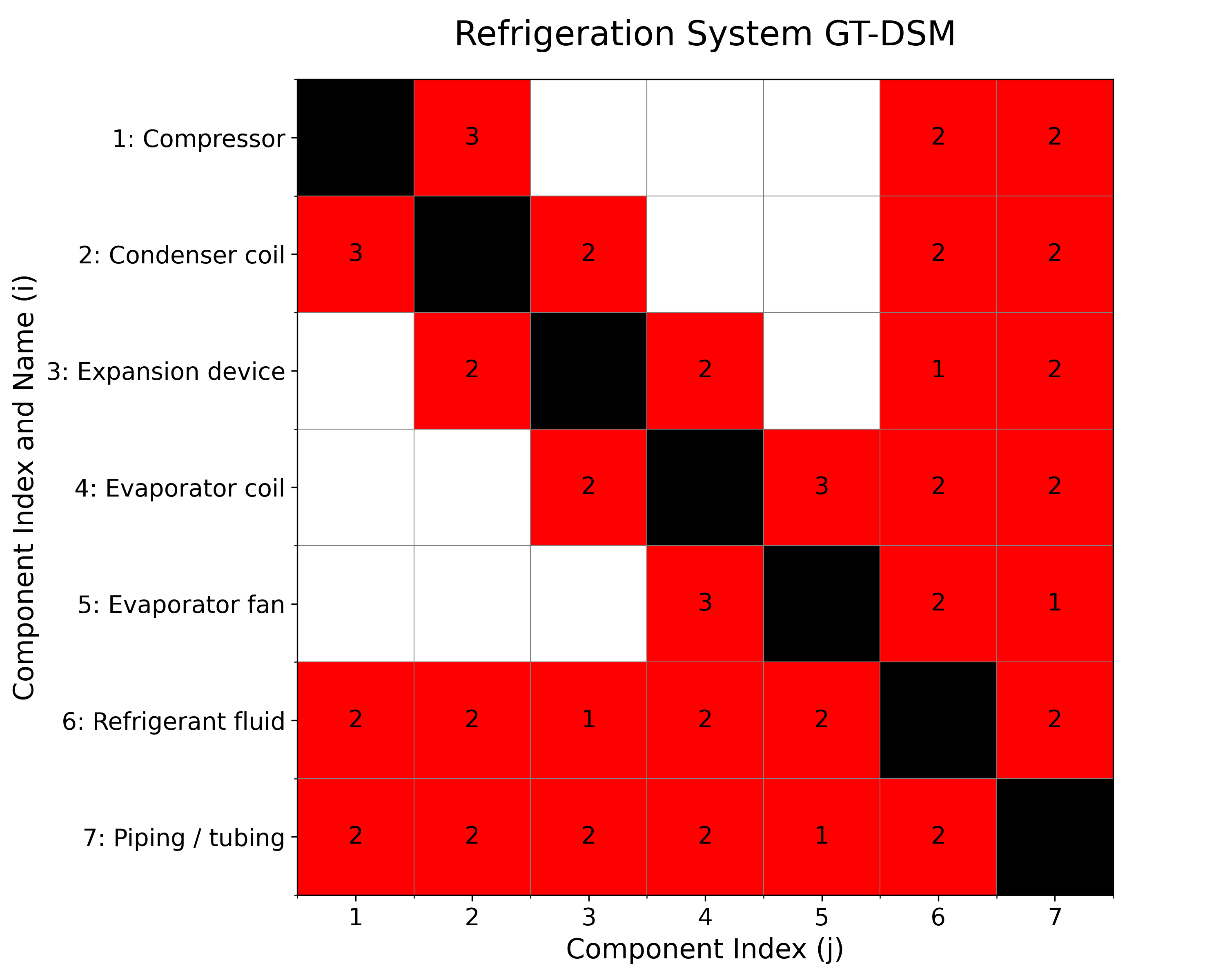}
    \caption{GT‑DSM of the Refrigeration System}
    \label{fig:GT-DSM:REF}
  \end{minipage}
\end{figure*}

Applying these strategies yields a structured and technically robust refrigerator decomposition. The ground truth dataset is manually validated for correctness and symmetry to ensure consistency. The system comprises 7 subsystems—\textit{Refrigeration System, Control System, User Interface \& Access System, Structural \& Insulation System, Defrost System, Lighting System, Drainage \& Moisture Control System}—each with, on average, more than 4 components, totaling 31 components. The dataset is stored in a spreadsheet with filterable columns and index numbers to support referencing when generating textual inputs. A partly preview is provided in \autoref{tab:excel_style_refrigerator}. The main statistics are summarized in Table \autoref{tab:decomposition_stats}, confirming the decomposition’s extensiveness and its suitability for comprehensive analysis by experimental customization.

\begin{table}[h]
\centering
\begin{tabular}{|l|r|}
\hline
\textbf{Metric} & \textbf{Value} \\
\hline
Subsystems & 7 \\
Components & 31 \\
Entries & 930 \\
Active links & 260 \\
Inactive links & 670 \\
Total \# Interactions & 450 \\
NZF & 28.92\% \\
Interaction types & 13 \\
\hline
\end{tabular}
\caption{System decomposition and DSM statistics}
\label{tab:decomposition_stats}
\end{table}

To evaluate metrics based on DSM characteristics and classifications, the dataset is manually converted into a GT-DSM and checked for correctness and symmetry to ensure consistency between the DSM and dataset. The resulting dataset and DSM forms a validated ground truth reference for assessing the quality and correctness of LLM-generated DSMs.

A more detailed overview of the refrigerator decomposition is provided in \autoref{tab:subsystem_metrics} of \autoref{Ap:Datasets} in the Appendix. Based on this overview, the \textit{Drainage \& Moisture Control} subsystem is selected as the low-complexity case due to its small size and low NZF for both inter- and intra-subsystem interactions; \autoref{fig:GT-DSM:D&M} also shows a modular structure supporting this choice. Among the remaining subsystems, the \textit{Refrigeration System} exhibits the highest complexity: it has the largest size and the highest inter- and intra-subsystem interconnectivity, and, as seen in \autoref{fig:GT-DSM:REF}, greater structural complexity than the Drainage \& Moisture Control system. A combined decomposition capturing both intra-subsystem and inter-subsystem interactions between the Refrigeration and Drainage \& Moisture Control subsystems is visualized in \autoref{fig:GT-DSM:REF+D&M} of \autoref{sec: Appendix A}.

\paragraph{Refrigerator input dataset}
For each experiment, technical documents for the relevant subsystem decompositions are generated from the ground truth. For the input parameter vs dataset experiments a controlled technical document is manually generated stating direct dependencies based on the generated GT-database to isolate semantic reasoning by matching the subsystem decomposition as much as possible to the fictive abstract technical document. However for complexity also more natural flowing documentation should be tested to see the semantic reasoning of the model. Using prompt engineering, the GT dataset is converted into textual technical documentation; the same strategies—Task Instruction Prompting, Constrained Generation, Chain-of-Thought Structuring, and format-forcing templates—are applied (see \autoref{sec: Appendix A}).

\begin{figure*}[t]
  \centering
  \includegraphics[width=0.8\textwidth]{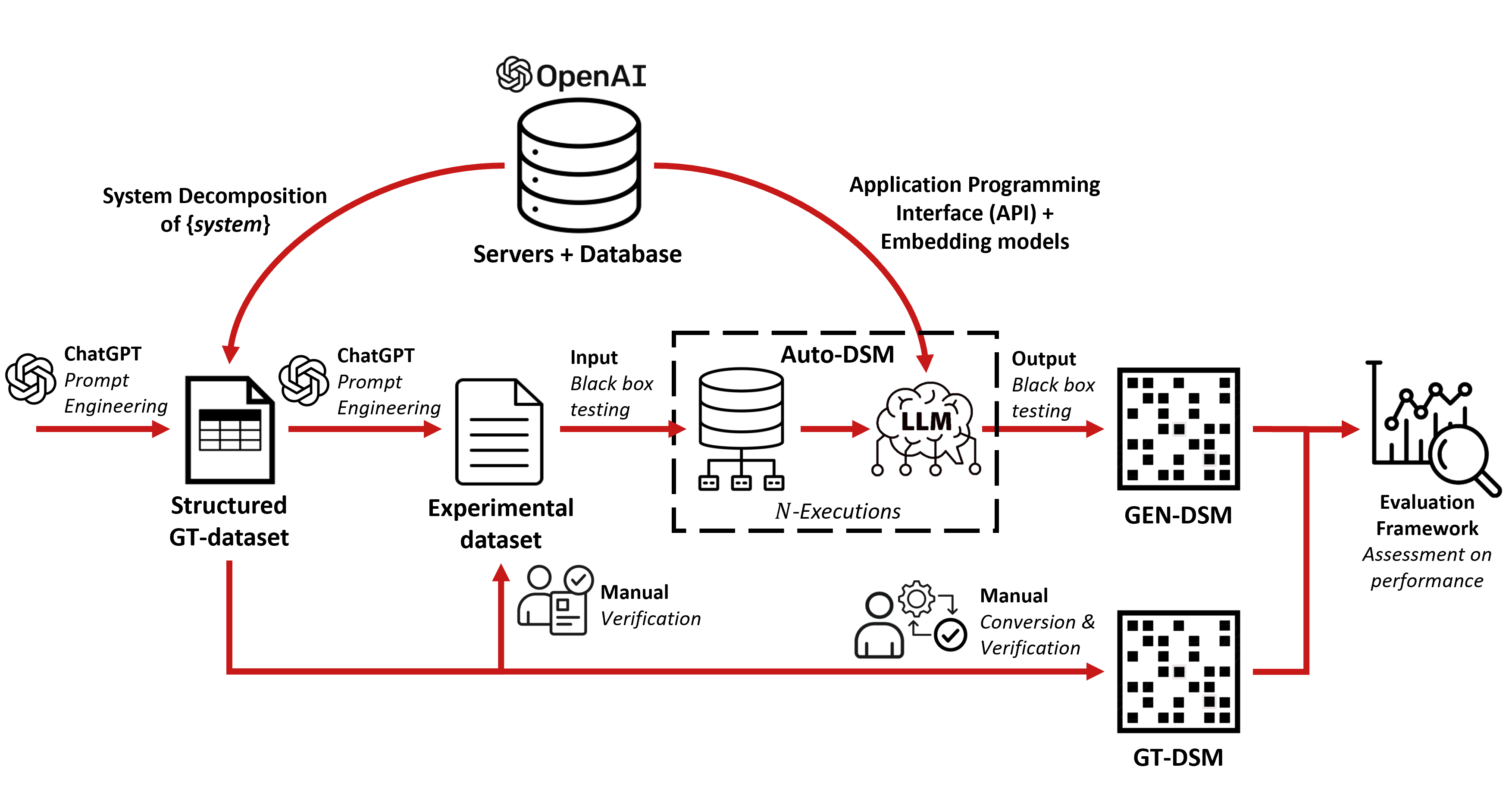}
  \caption{Schematic Overview of the Refrigerator System Decomposition Experimental Setup}
  \label{fig:dataflow_experiment}
\end{figure*}

To construct experimental datasets, the corresponding subsystems and interactions are first filtered from the GT dataset. This selection is inserted into the prompt and executed in ChatGPT (GPT-4o). The prompt in \autoref{lst:experprompt1} focuses on phrasing the GT data for a provided list of inter-subsystem interactions. To guarantee alignment between the GT dataset and the technical document, each statement includes an index number for reference. This facilitates manual verification and correction of any hallucinations or inconsistencies, yielding a well-defined, GT-consistent dataset.

The overall data-generation and execution process for the refrigerator experiments is shown in \autoref{fig:dataflow_experiment}. This flow is also representable for manual generated technical documents, however then the process step between "\textbf{Structured GT-dataset}" and "\textbf{Experimental dataset}" should be "\textbf{Human}" \textit{Manually generated} instead of "\textbf{ChatGPT}" \textit{Prompt Engineering}.

%% file: 5_Results_Analysis/Results_Analysis.tex
\section{Results \& Analysis}
\begin{table*}[t]
\centering
\caption{Baseline Agreement and Reproducibility Metrics for Different Input Parameter Sets}
\label{tab:agreement_results}
\begin{tabular}{l l c c c}
\toprule
\textbf{Baseline Dataset} & \textbf{Input Parameter Set} & \textbf{N} & \textbf{Agreement Rate (Mean $\pm$ SD)} & \textbf{Fleiss' $\kappa$} \\
\midrule
One directional abstract & Abstract; - & 30 & 0.91 $\pm$ 0.16 & 0.78 \\
Dependency type abstract & Abstract; - & 30 & 0.99 $\pm$ 0.04 & 0.97 \\
Drainage \& Moisture Control System & Refrigeration System; Alignment & 30 & 1.00 $\pm$ 0.00 & 1.00 \\
Refrigeration System & Refrigeration System; Gas Material & 30 & 0.96 $\pm$ 0.11 & 0.86 \\
\bottomrule
\end{tabular}
\end{table*}

\autoref{tab:agreement_results} summarizes the reproducibility statistics for four representative baseline datasets evaluated using the Auto-DSM pipeline. Each dataset was tested using a consistent input parameter configuration over $N=30$ independent executions. The raw agreement rate (mean and standard deviation) and Fleiss’ $\kappa$ are reported to characterize both consistency and chance-corrected reproducibility.

All four cases exhibit strong reproducibility signals, with mean agreement rates exceeding 0.90 and $\bar{\kappa}$ values well above the conservative threshold of 0.60. These results empirically validate the sufficiency of the initial sample size $N = 30$ in all tested configurations. 

\subsection{Interaction Definition}
The datasets containing different interaction definitions are compared to the GT-DSM as well as to each other to gain insight into the classification behavior of the Auto-DSM. While this will provide a first estimation on the influence of semantic reasoning within the technical documentation by the pipeline, the results are not comprehensive enough and will be supported by other experiments to come to a conclusion on the classification limits, behavior and performance.

\subsubsection{Uni-directional Dependency Definition}
\begin{figure}[t]
    \centering
    \includegraphics[width=\linewidth]{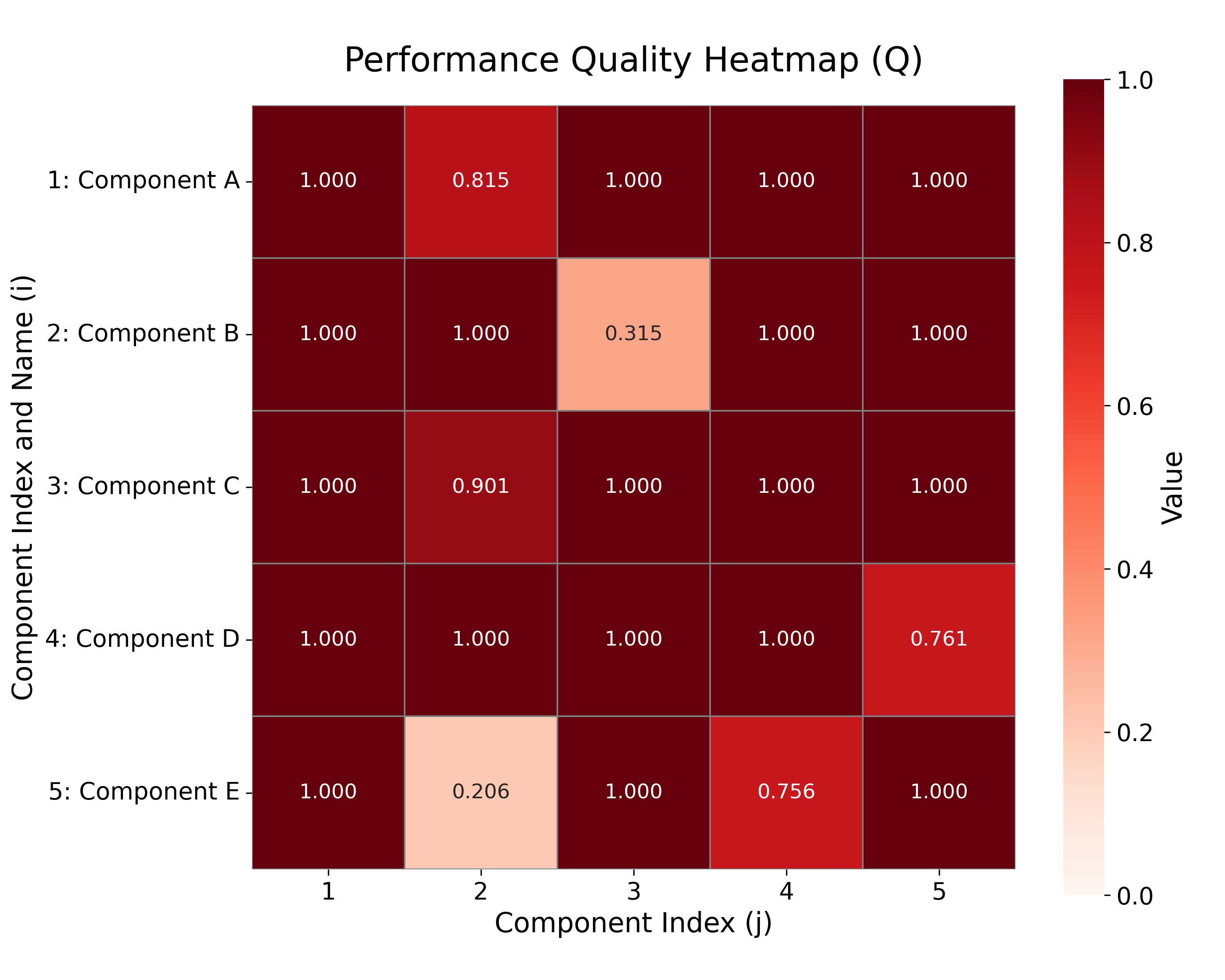}
    \caption{Performance Quality Heatmap of the Uni-directional Dependency Definition for an Abstract System}
    \label{fig:Q-Uni-Abst}
\end{figure}

The first experiment contains the baseline experiment with the dataset of the non-symmetric abstract system decomposition between components A-E. The notation for every cell in the DSM is stated as $RC$, where $R= Row, C = Column$. The per cell level Quality performance is visualized in \autoref{fig:Q-Uni-Abst}, indicating 6 problematic cells for the dependency definitions: $AB, BC, CB, DE, EB, ED$. A first observation of \autoref{fig:Q-Uni-Abst} indicates that the performance of the bottom left triangle is better than the top right triangle. Indicating that directional asymmetry is systematically present. Since Auto-DSM executes the dependency type identification based per column, it first searches for the column to row dependency and afterwards to the row to column dependency. Looking at the dependency between A and B this is noted as: "Component A is linked to Component B", filling in column A row B first and Column B row A second with the defined prompt resulting in the prompt question direction $A\rightarrow B$ where $Q_{BA} = 1$ while $B\rightarrow A$ results in $Q_{AB} = 0.81$. This is also the case for the dependency between component B and component C. While this seems like a reasonable conclusion, the other two pairs do not support this hypothesis since $Q_{ED}=Q_{DE}$ and $Q_{BE}>Q_{EC}$. In order to make a conclusion the interaction definitions should be investigated to see if this influences the Auto-DSM responses. 

\begin{table}[h!]
\centering 
\begin{tabular}{l c c c c c c}
\hline
\textbf{Cell} & \textbf{GT} & $H_{\text{norm}}$ & $p(-1)$ & $p(0)$ & $p(1)$ & \textbf{$M_{rc}$} \\
\hline
A$\rightarrow$B & 1  & 0.53 & 0.00 & 0.27 & 0.73 & 0.73 \\
B$\rightarrow$C & 1  & 0.49 & 0.77 & 0.00 & 0.23 & 0.23 \\
C$\rightarrow$B & 1  & 0.30 & 0.00 & 0.10 & 0.90 & 0.90 \\
D$\rightarrow$E & 1  & 0.63 & 0.00 & 0.50 & 0.50 & 0.50 \\
E$\rightarrow$B & $-1$ & 0.70 & 0.03 & 0.63 & 0.34 & 0.03 \\
E$\rightarrow$D & 1  & 0.61 & 0.00 & 0.60 & 0.40 & 0.40 \\
\hline
\end{tabular}
\caption{Non-zero entropy values, the probabilities for $N$ responses and the matched frequency based on the Ground Truth value (GT).}
\label{tab:entropyuni}
\end{table}

Looking into the entropy matrix the exact same 6 cells indicate different levels of variability by being non-zero for entropy. In order to understand the behavior of these cells and the dependencies, the normalized entropy is divided into its response probabilities, and compared to the GT-DSM. I.e. looking into the matched, uncertainty and mismatched frequencies, allowing for a structured overview presented in \autoref{tab:entropyuni}. 

A first observation is that Auto-DSM has uncertainty in its confidence to assign a dependency or rejects the dependency classification. For cell AB, CB, DE and ED Auto-DSM is inconsistent in its confidence to identify a dependency. However, if Auto-DSM responds with confidence the dependency identification is correct. Looking into the dependency definition for these cases, these are all direct defined dependencies and direct defined dependencies only. Indicating that it could be correlated to the interaction definition. However, cell BC is also a direct defined dependencies but this cell has no problems with confidence but repeatedly classifies the dependency incorrectly as an absent dependency (77\%), contradicting the confidence uncertainty pattern. While all direct dependencies are stated the same, the classification performance variates largely from matching 100\%, to 40\% with an uncertainty rate of 60\%, to 100\% confident response with 77\% incorrect responses indicating variability introduced by other factors. Cell EB is particularly interesting, since it is one of the 10 undefined dependency definitions, where every other undefined dependency definition is identified correctly. EB not only shows a variability in confidence but also highlighting contradictions within the confident classifications. This stochastic behavior would be explainable to the dependency not being defined, however one would expect to also see this behavior for the other 9 undefined dependencies, indicating again high but unexplainable variability based on interaction definition. 

Looking into the dependency definitions over the complete DSM and not only the non-perfect quality scores, the following insights are gathered: Auto-DSM predicts the direct dependencies for 62,7\% correct, 12,8\% incorrect and 24,5\% of the times rejecting the classification. Indirect dependencies as well as absent dependencies are 100\% predicted correctly by as absent. Undefined dependencies are predicted 90,3\% of the times correct, 3,4\% of the times incorrect and 6,3\% of the time rejected. Where the technical document gives clear direct evidence of different type of dependency definitions, we still see splits between correct, uncertain or even incorrect responses. These splits could be coming from definition ambiguity in the prompt by leaving the linkage-type empty or the direction not being specified allowing for hallucinations and freedom of interpretation of the model. So while the dependency definition may have somewhat of an influence on the dependency identification, there is not enough evidence to make a clear and direct correlation between performance and different types of uni-directional interaction definitions.

\begin{figure}[t]
    \centering
    \includegraphics[width=\linewidth]{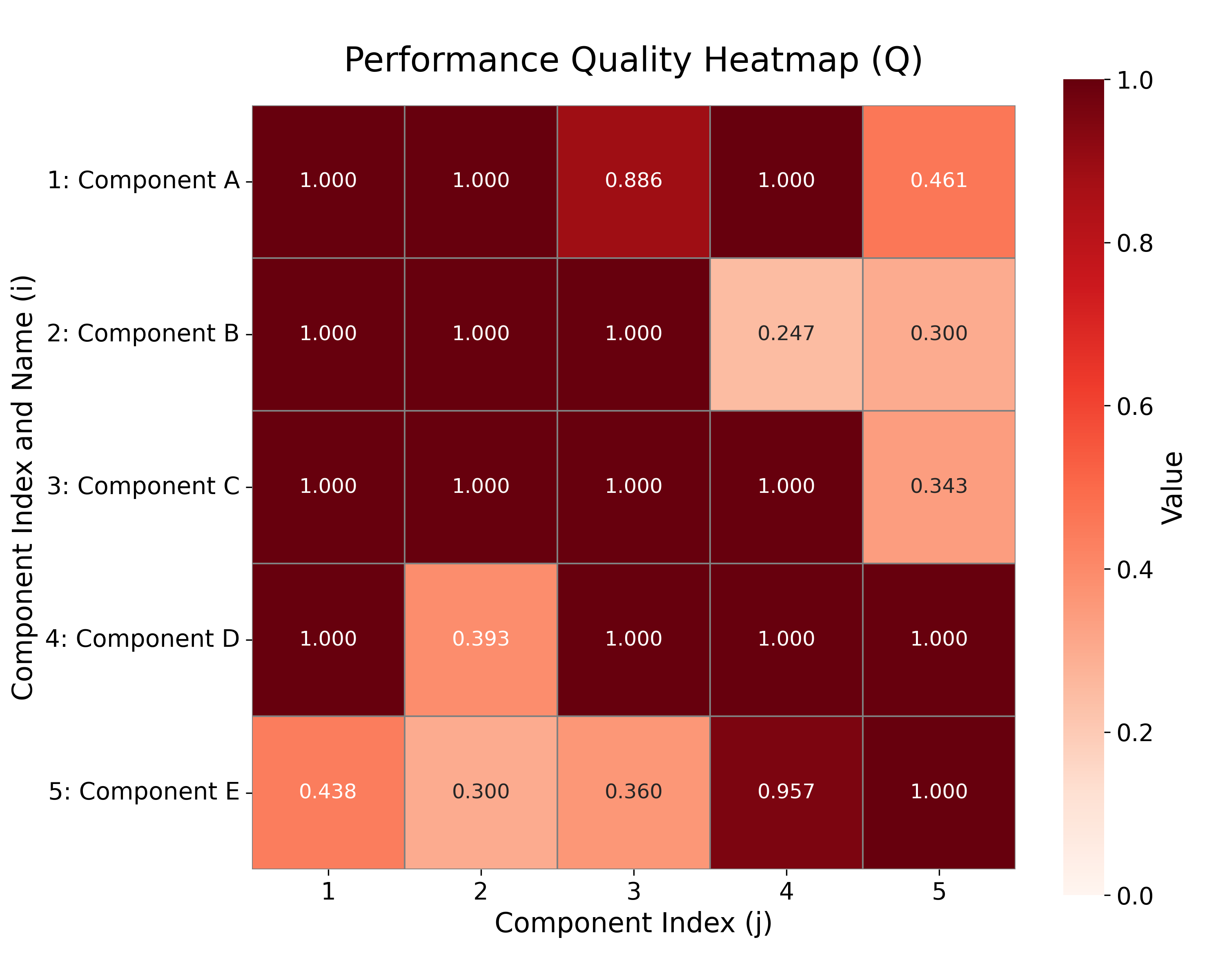}
    \caption{Performance Quality Heatmap of the Bi-directional Dependency Definition for an Abstract System}
    \label{fig:Q_ID_BI}
\end{figure}

\subsubsection{Bi-directional Dependency Definition}
When incorporating a bidirectional statements of the system decomposition for the same sample size and dataset, there is a remarkable difference for the pattern in the performance metric Q heatmap visualized in \autoref{fig:Q_ID_BI} compared to  Uni-directional. Introducing Bi-directional dependency definitions improved the performance for direct, indirect and absent defined dependencies. This pattern is expected and indicates correct reasoning of the Auto-DSM method, since defined dependencies should be confidently predicted while undefined dependencies introduce ambiguity in data input where uncertainty in responses is expected to increase. Comparing the bottom left triangle to the top right triangle of the performance metric highlights the increase in symmetry performance, where this was pattern was less visible for the uni-directional abstract system. This is backed by the average Correctness parameter for both cases, where the correctness for Uni-directional is on average 46,16\% with a $\sigma=30\%$, the Bi-directional clearly indicates a increase with an average of 95,93\% and a $\sigma = 4,63\%$. Cell AC and ED are scoring the lowest on the performance indicator with a score of 0.886 and 0.957, looking into the entropy and its probabilities compared to the GT-DSM cell AC was predicted incorrectly less than 7\% of the runs, while ED was predicted as uncertain even less only 3\%. This error margin is lower than the 90\% agreement rate found by \cite{Schmitz2011} but the complexity of the system decomposition is not comparable between cases, Auto-DSM has to be incorporated in order to assess if the classification performance stays within the range for more complex tasks. 

Secondly, it is remarkable that with the uni-directional it seemed that Auto-DSM predicted confidently undefined dependencies as an absence, the Bi-directional however has the same confidence with a highest uncertainty percentage not exceeding 3\% but the predictions made are largely incorrect ranging within the incorrect prediction percentages between 50\% and even 100\% for non-perfect quality scores. Both the cells EB and BE predicted even 100\% incorrectness, which is remarkable since this was also a problematic cell for the uni-directional system decomposition. This is unrelated to the dependency definition since these findings only apply to this specific pair, but indicate confident repeatable incorrect prediction behavior. This could be due to the lack of defining a dependency type  in the input parameter setting in combination with the presence of undefined dependency definitions in the input dataset. This combination introduces a lot of ambiguity, where a model could fall back to its trainings data recognizing patterns or knowledge instead of rejecting the classification as also discussed by Li et al. \cite{Li2025}. To test this experiments between the input parameter and the dataset should be preformed.

\subsubsection{Inconsistent Dependency Definitions}
Both the documents for the duplication statement and Contra-dictionary statement are executed with the sample size defined by the baseline execution. Since the non-symmetric DSM provided insights in high uncertainty for the dependency pair DE and ED, this pair is selected to see if introducing duplications has influence on the classification predictions. The duplication performance Q heatmap has the same pattern for the direct, indirect and absence dependencies between components A, B, C and D as for the single stated Uni-directional experiment, indicating the same prediction behavior but with some small deviations due to the uncertainty behavior in confidence. Undefined dependency definitions highlight again stochastic behavior ranging from a perfect quality score of 1 to a $Q_{EB}=0.165$, again for cell EB. However, for the dependency pair DE/ED the performance went up to a perfect classification quality score of 1 visualized in \autoref{tab:Q_Incon}. This proves  the statement Koh made on pooling is indeed true for Auto-DSM \cite{Koh2024}, where more of the same data contributes to a more confident response. 

\begin{table}[h!]
\centering 
\setlength{\tabcolsep}{4.5pt} 
\begin{tabular}{l l c c c c c c c}
\hline
\multirow{2}{*}[-0.2em]{\shortstack{\textbf{Experiment}}} 
 & \multirow{2}{*}[-0.2em]{\shortstack{\textbf{Cell} \\ (R$\rightarrow$C)}} 
 & \multirow{2}{*}[-0.2em]{\textbf{GT}} 
 & \multirow{2}{*}[-0.2em]{$Q_{\text{n}}$} 
 & \multirow{2}{*}[-0.2em]{$H_{\text{n}}$} 
 & \multirow{2}{*}[-0.2em]{$p(-1)$} 
 & \multirow{2}{*}[-0.2em]{$p(0)$} 
 & \multirow{2}{*}[-0.2em]{$p(1)$} 
 & \multirow{2}{*}[-0.2em]{$M_{rc}$} \\
 & & & & & & & & \\
\hline
\multirow{2}{*}{\shortstack{Duplication}} 
 & D$\rightarrow$E & 1 & 1.00 & 0.00 & 0.00 & 0.00 & 1.00 & 1.00 \\
 & E$\rightarrow$D & 1 & 1.00 & 0.00 & 0.00 & 0.00 & 1.00 & 1.00 \\
\hline
\multirow{4}{*}{\shortstack{Single Uni-\\Directional}} 
 & D$\rightarrow$E & 1 & 0.76 & 0.63 & 0.00 & 0.50 & 0.50 & 0.50 \\
 & E$\rightarrow$D & 1 & 0.75 & 0.61 & 0.00 & 0.60 & 0.40 & 0.40 \\
 & A$\rightarrow$B & 1 & 0.82 & 0.53 & 0.00 & 0.27 & 0.73 & 0.73 \\
 & B$\rightarrow$A & 1 & 1.00 & 0.00 & 0.00 & 0.00 & 1.00 & 1.00 \\
\hline
\multirow{2}{*}{\shortstack{Present $\rightarrow$\\ Absence}} 
 & A$\rightarrow$B & 0 & 0.30 & 0.00 & 1.00 & 0.00 & 0.00 & 0.00 \\
 & B$\rightarrow$A & 0 & 0.30 & 0.00 & 1.00 & 0.00 & 0.00 & 0.00 \\
\hline
\multirow{2}{*}{\shortstack{Absence $\rightarrow$\\ Present}} 
 & A$\rightarrow$B & 0 & 0.30 & 0.13 & 0.00 & 0.03 & 0.97 & 0.03 \\
 & B$\rightarrow$A & 0 & 0.28 & 0.00 & 0.00 & 0.00 & 1.00 & 0.00 \\
\hline
\end{tabular}
\caption{Probabilities $p(i)$, normalized metrics $Q_n$ and $H_n$, and matched rate $M_{rc}$ values based on Ground Truth (GT) for inconsistency dependency definition experiments.}
\label{tab:Q_Incon}
\end{table}

Next to the duplication experiment, also contradicting dependency definitions were tested to identify the ability of Auto-DSM to detect problematic dependencies and classifying them as an uncertainty by rejection. Contradicting dependency definitions are tested on the best performing dependency pair between A and B of the single statement Uni-directional experiment. In order to eliminate the effect of statement sequence two experiments are tested, direct dependency followed by an absence dependency and vice versa. \autoref{tab:Q_Incon} shows that the performance Q, the entropy and the probabilities compared to the GT-DSM are heavily influenced by not only the contradiction but also by the sequence of the contradiction. Expected would be that at the uncertainty would increase, however the results in \autoref{tab:Q_Incon} show that the predictions are inclined to confident but incorrect predictions to the last stated dependency definition in the sequence instead of rejecting due to ambiguity in description. 

The experiments demonstrate that interaction definition exerts a measurable yet inconsistent influence on Auto-DSM’s classification performance. Uni-directional dependency definitions reveal directional asymmetry and high variability in confidence and correctness, with problematic cells (e.g., EB, BC) exhibiting stochastic misclassification patterns that cannot be fully attributed to the interaction type alone. Bi-directional definitions substantially improve correctness, reduce variance, and enhance symmetry across the DSM, yet also introduce instances of confidently incorrect predictions, particularly for undefined dependencies. Inconsistent definitions highlight Auto-DSM’s sensitivity to redundancy and sequence effects, where duplication increases confidence and correctness, but contradictory inputs often result in confident misclassifications rather than uncertainty. Collectively, these findings indicate that while clear and consistent interaction definitions improve performance, Auto-DSM remains susceptible to ambiguity-induced bias and systematic errors, necessitating further controlled experiments to isolate the contribution of interaction semantics from other influencing factors.

\begin{table*}[h!]
\centering
\setlength{\tabcolsep}{4.5pt}
\renewcommand{\arraystretch}{1.2}
\begin{tabular}{l l l l c c cc cc cc cc}
\hline
\multirow{2}{*}{\textbf{Exp.}}
 & \multirow{2}{*}{\textbf{System}}
 & \multirow{2}{*}{\shortstack{\textbf{Dependency}\\\textbf{Type}}}
 & \multirow{2}{*}{\textbf{Dataset}}
 & \multirow{2}{*}{\shortstack{\textbf{Comp.}\\\textbf{Mode}}}
 & \multirow{2}{*}{\shortstack{\textbf{Occ.}\\\textbf{(\%)}}}
 & \multicolumn{2}{c}{\boldmath$Q_{\text{norm}}$}
 & \multicolumn{2}{c}{\boldmath$H_{\text{norm}}$}
 & \multicolumn{2}{c}{\textbf{SA}}
 & \multicolumn{2}{c}{\textbf{Completeness}} \\
 & & & & & & \textbf{Mean} & \textbf{Std} & \textbf{Mean} & \textbf{Std} & \textbf{Mean} & \textbf{Std} & \textbf{Mean} & \textbf{Std} \\
\hline
1.1 & Abstract     & Mechanical          & Bi-direc. incl. [ME, TE]   & 5         & 100    & 0.99 & 0.03 & 0.02 & 0.08 & 1.00 & 0.00 & 99.17 & 1.86 \\
2.1 & Abstract     & Electrically        & Bi-direc. incl. [ME, TE]   & 5         & 100    & 0.71 & 0.23 & 0.22 & 0.25 & 1.00 & 0.00 & 14.33 & 13.46 \\
3.1 & Abstract     & --                  & Bi-direc. incl. [ME, TE]   & 5         & 100    & 0.88 & 0.22 & 0.11 & 0.18 & 0.84 & 0.05 & 99.33 & 1.70 \\
4.1 & Abstract     & --                  & Bi-direc. excl. IT         & 5         & 100    & 1.00 & 0.00 & 0.00 & 0.00 & 1.00 & 0.00 & 100   & 0.00 \\
1.2 & D\&M         & Mixture Material      & D\&M                       & 4         & 100    & 0.81 & 0.29 & 0.09 & 0.26 & 0.69 & 0.04 & 97.50 & 3.82 \\
2.2 & D\&M         & Mechanical          & D\&M                       & 4         & 100    & 0.72 & 0.33 & 0.05 & 0.15 & 0.47 & 0.03 & 96.11 & 4.16 \\
3.2 & D\&M         & --                  & D\&M                       & 4         & 100    & 0.81 & 0.29 & 0.09 & 0.20 & 0.69 & 0.07 & 99.72 & 1.50 \\
4.2 & D\&M         & --                  & D\&M excl. IT              & 4         & 100    & 0.48 & 0.30 & 0.00 & 0.00 & 0.00 & 0.00 & 100   & 0.00 \\
5--8 & Bicycle     & Mixture Material      & D\&M                       & 7, 8, 9   & 77--97 & N/A  & N/A  & N/A  & N/A  & N/A  & N/A  & N/A   & N/A \\
9   & Abstract     & Mixture Material      & Bi-direc. incl. [MM,A, P]  & 5         & 100    & 0.91 & 0.15 & 0.17 & 0.25 & 0.95 & 0.04 & 83.33 & 5.68 \\
10  & D\&M         & Proximity           & D\&M                       & 4         & 100    & 0.74 & 0.34 & 0.00 & 0.00 & 0.50 & 0.00 & 100   & 0.00 \\
11  & D\&M         & Alignment           & D\&M                       & 4         & 100    & 1.00 & 0.00 & 0.00 & 0.00 & 1.00 & 0.00 & 100   & 0.00 \\
12  & D\&M         & Thermally           & D\&M                       & 4         & 100    & 0.69 & 0.27 & 0.10 & 0.19 & 1.00 & 0.00 & 15.28 & 4.85 \\
13  & Bicycle      & Mixture Material      & D\&M incl. Bicycle         & 4         & 100    & 0.84 & 0.28 & 0.08 & 0.21 & 0.74 & 0.05 & 100   & 0.00 \\
14  & Bicycle      & --                  & D\&M incl. Bicycle         & 4         & 100    & 0.87 & 0.26 & 0.05 & 0.15 & 0.77 & 0.04 & 100   & 0.00 \\
15  & D\&M         & Mixture Material      & D\&M excl. Comp. Sum       & 4         & 100    & 0.76 & 0.32 & 0.03 & 0.12 & 0.56 & 0.04 & 100   & 0.00 \\
16  & Refrigerator & Mixture Material      & D\&M                       & 4         & 100    & 0.81 & 0.29 & 0.09 & 0.26 & 0.69 & 0.04 & 97.50 & 3.82 \\
17  & Refrigerator & --                  & D\&M                       & 4         & 100    & 0.81 & 0.29 & 0.09 & 0.20 & 0.69 & 0.07 & 99.72 & 1.50 \\
18--19 & Refrigerator & Mixture Material, -- & D\&M excl. Refrigerator   & IDK       & 57--87 & N/A  & N/A  & N/A  & N/A  & N/A  & N/A  & N/A   & N/A \\
\hline
\end{tabular}
\caption{Overview of experiments with component modes, occurrence, normalized quality and entropy metrics ($Q_{\text{norm}}$, $H_{\text{norm}}$), Selective Accuracy (SA), and Completeness and the metrics mean and standard deviations.}
\label{tab:experiments_full}
\end{table*}

\subsection{System Parameters vs Dataset}
The \textit{System Parameters vs Dataset} experiments were executed using (i) the refined bi-directional abstract system decomposition with interaction types and (ii) the refrigerator subsystem \textit{Drainage \& Moisture Control} (D\&M). For this analysis, both manually generated technical documents based on the abstract decomposition and the refrigerator GT-dataset were created to state only direct, symmetric dependencies. This isolates the influence of system-parameter selection from the dataset by excluding added complexity and leveraging results from the dependency-definition experiments. Building on the interaction-definition experiments, DSM-wide aggregated performance metrics were used for analysis.

As explained in the Experimental Setup, eight experiments were executed. These are divided into two categories by objective: Experiments 1–4 analyze the abstention behavior of Auto-DSM in dependency identification, while Experiments 5–8 focus on abstention in system component identification. Observed inaccuracies and inconsistencies motivated follow-up experiments to diagnose problematic responses. \autoref{tab:experiments_full} reports the results for $N=30$, listing the input parameter, input dataset, the most frequently classified number of components (Component Mode), the occurrence of this mode, and the mean and standard deviation of performance quality, entropy, selective accuracy, and completeness, to characterize accuracy, variability, and uncertainty DSM-wide relative to the GT.

\subsubsection{Dependency Type Parameter vs Dataset}
The first four experiments assess the influence of input parameters versus dataset for dependency-type interactions, examining behavior when uncertainty is introduced by increased strictness or by mismatches between input parameters and dataset.

\paragraph{Abstract system}
For the abstract system, direct dependencies were refined by adding a dependency type. Utilizing insights from the interaction-definition experiments, we evaluated overall DSM performance and the effect of dependency types when pattern and definition influences are known. Experiment 1.1 produced an almost perfect score, except for cell AB (direct mechanical statement), which showed an uncertainty frequency of 0.17. Experiment 2.1 yielded a lower performance score and very low completeness: the model preferred abstention over asserting absence, even though the document clearly contained no electrical dependency types; when confident, classifications were perfect. Experiment 3.1 mirrored the dependency-definition results: high performance for direct and absent links, with lowest match frequency 0.9, and all mismatches due to uncertainty responses. Undefined dependencies exhibited substantial uncertainty and inaccuracy when the model was confident, increasing entropy and lowering selective accuracy. Experiment 4.1 produced a consistent all-zeros DSM, correctly acknowledging knowledge boundaries.

\paragraph{Real system}
Guided by these findings, a real system was evaluated using a manually generated dataset for the D\&M subsystem, represented in \autoref{lst:ManM&D.txt}, which contains only bi-directional direct statements. As expected, Experiments 1.2 and 3.2 provided roughly the same performance because every dependency pair included the \textit{Mixture Material} interaction type. However, relative to the abstract findings, near-perfect scores were expected (as in Experiment 1.1), since dependencies were directly stated and included in the dataset. Three explanations are plausible: (i) real-system effects, (ii) poor interpretation of the dependency type by the LLM, and (iii) overgeneralization because all direct dependencies in the technical document share the same input dependency type.

To verify whether the dependency-type parameter caused hallucinations—potentially because Auto-DSM recognizes the system and defaults to pretrained knowledge—Experiment 9 swapped the abstract system’s “mechanically” dependencies for “Mixture Material,” yielding an identical dataset but with D\&M dependency types present. This resulted in a high performance score, $Q_n=0.91$, 0.08 lower than Experiment 1.1 but higher than Experiment 1.2. Entropy and completeness increased relative to Experiment 1.1, suggesting that the LLM has more difficulty with this HDDSM dependency type, although it is not the sole factor in the unexpected behavior. The model exhibited increased uncertainty by abstaining on some classifications, yet the results indicate the hallucination was not introduced by the dependency type and that the HDDSM framework can be classified \cite{Tilstra2012}. Note that \textit{Mixture Material} is defined for every direct dependency in the D\&M dataset, which may cause overgeneralization.

To investigate overgeneralization further, Experiments 10 and 11 changed the input parameter to other dependency types. Experiment 10 showed worse performance, $Q_n=0.74$, despite $H_n=0$ and 100\% completeness: it confidently misclassified the Drain Pan–Drain hole pair as Proximity and classified the Drain pan–evaporation tray pair—expected to be Proximity—as having no Proximity dependency. These results contradict the direct definitions in \autoref{lst:ManM&D.txt} and reveal a previously unobserved hallucination under direct definitions, potentially caused by the model defaulting to pretrained knowledge when prompts are weakly constrained. Experiment 11 checked the third dependency type; as shown in \autoref{tab:experiments_full}, the dependency-type identification was completely correct and deterministic, aligning with the abstract-system experiments. This fluctuation in system-type parameter identification may arise because the Tilstra framework has not been benchmarked for LLM-based dependency classification; semantic reasoning may allow the LLM to identify some dependency types more reliably than others. Overgeneralization is also plausible: when all defined dependencies share the same type, entropy and completeness appear strong while selective accuracy drops, yielding an almost all-ones GEN-DSM. Exploring other or larger systems may clarify whether, when unsure, the LLM recognizes systems and falls back to training data rather than acknowledging knowledge boundaries. 

Concluding for these dependency-type experiments: in real-world systems, dependency types combined with real components tend to trigger learned patterns or pretrained knowledge (Experiments 2.2 and 4.2), whereas abstract systems (Experiments 2.1 and 4.1) better acknowledge knowledge boundaries. In 2.1 and 4.1, high uncertainty responses drove completeness toward zero; in 2.2 and 4.2, the model was confident but incorrect. Experiment 4.2 notably returned an all-present GEN-DSM where a full-uncertainty matrix (as in 4.1) was expected. Abstract-system experiments provided consistently higher performance, and the mechanical interaction category scored slightly better than \textit{Mixture Material}, indicating that the LLM can reason with the Tilstra HDDSM framework but recognizes some interaction categories more reliably than others. 

\subsubsection{System Definition Mismatch (Component Identification)}
Where Experiments 1–4 (and follow-ups 9–11) exposed inconsistencies for the dependency input parameter, the system-definition mismatch reveals issues in component identification by Auto-DSM. In Experiments 5–8, the concise D\&M decomposition served as the dataset and the input system parameter was \textit{Bicycle}. For $N=30$, more than 75\% of responses identified real bicycle components such as \textit{Pedals} or \textit{Chain}. For \textit{Mixture Material}, the dependency analysis for the non-rejected component sets produced an all-zeros matrix, correctly indicating knowledge boundaries for inter-component dependencies. In contrast, for \textit{Mechanically}, the model confidently classified some bicycle components as having no dependency, revealing a chain of hallucinations and a tendency to respond confidently rather than abstain.

To gain more insight into component-identification hallucinations, Experiments 13 and 14 tested whether Auto-DSM prefers pretrained knowledge over factually correct but real-world-incongruent decompositions. The input parameter remained \textit{Bicycle} with the DSM decomposition, and the document opened with:
\begin{quote}
\textit{"A bicycle is developed, containing the following components: The drain hole, drain pan, evaporation tray, and moisture control filters, which interact through the below stated dependencies to ensure functionality."}
\end{quote}
Both cases produced results similar to Experiments 1.2 and 3.2, indicating that component identification prioritized the technical documentation rather than defaulting to training data or rejecting due to conflict. Experiment 15 explored whether removing the initial component list affects identification; as shown in \autoref{tab:experiments_full}, the 4/4 components occurred with 100\% frequency for $N=30$, indicating no effect. Experiments 16–19 examined whether pretrained knowledge fills uncertainty in real-system decompositions. First, when the document stated that D\&M is a refrigerator subsystem (input parameter: \textit{refrigerator}; Experiments 16 and 17), results matched those of 1.2 and 3.2, confirming correct hierarchical identification. For Experiments 18 and 19, this statement was omitted, introducing ambiguity resulting in the abstention frequency 57-87\%. When the subsystem appeared in the dataset, Auto-DSM correctly identified all components. However, once the hierarchical structure was excluded from the technical document, the LLM again showed inconsistent recognition of knowledge boundaries (as with the bicycle example), with more frequent abstention and high variance in confidence, indicating instability.

This highlights that Auto-DSM prioritizes information in the technical documentation when it matches the input parameter. In the absence of such alignment, the LLM defaults to pretrained knowledge and hallucinates by introducing random components instead of issuing uncertain rejections.

To conclude the findings for the input-parameters–dataset experiments, responses exhibit inconsistency and inaccuracy. While these experiments set the stage, they already underscore the importance of evaluating such pipelines across multiple categories. The findings relate to \cite{Li2025}, which suggests that an LLM struggles to acknowledge its boundaries. Once the technical document provides a well-defined system decomposition—whether aligned with its training data or not—Auto-DSM can identify components.

\begin{table*}[h!]
\centering
\setlength{\tabcolsep}{4.5pt}
\renewcommand{\arraystretch}{1.2}
\begin{tabular}{l l l l c c cc cc cc cc}
\hline
\multirow{2}{*}{\textbf{Exp.}}
 & \multirow{2}{*}{\textbf{System}}
 & \multirow{2}{*}{\shortstack{\textbf{Dependency}\\\textbf{Type}}}
 & \multirow{2}{*}{\textbf{Dataset}}
 & \multirow{2}{*}{\shortstack{\textbf{Comp.}\\\textbf{Mode}}}
 & \multirow{2}{*}{\shortstack{\textbf{Occ.}\\\textbf{(\%)}}}
 & \multicolumn{2}{c}{\boldmath$Q_{\text{norm}}$}
 & \multicolumn{2}{c}{\boldmath$H_{\text{norm}}$}
 & \multicolumn{2}{c}{\textbf{SA}}
 & \multicolumn{2}{c}{\textbf{Completeness}} \\
 & & & & & & \textbf{Mean} & \textbf{Std} & \textbf{Mean} & \textbf{Std} & \textbf{Mean} & \textbf{Std} & \textbf{Mean} & \textbf{Std} \\
\hline
1.2  & D\&M & Mixture Material & D\&M & 4 & 100 & 0.81 & 0.29 & 0.09 & 0.26 & 0.69 & 0.04 & 97.50 & 3.82 \\
20   & REF  & Liquid Material & REF  & 7 & 100 & 0.84 & 0.28 & 0.07 & 0.19 & 0.77 & 0.02 & 95.40 & 2.03 \\
21   & D\&M & Gas Material    & D\&M & 4 & 100 & 0.83 & 0.28 & 0.05 & 0.15 & 0.71 & 0.04 & 100   & 0.00 \\
22   & REF  & Gas Material    & REF  & 7 & 100 & 0.93 & 0.21 & 0.01 & 0.09 & 0.89 & 0.01 & 100   & 0.00 \\
3.2 & D\&M & -- & D\&M & 4 & 100 & 0.81 & 0.29 & 0.09 & 0.20 & 0.69 & 0.07 & 99.72 & 1.50 \\
23 & D\&M & -- & D\&M Extensive & 4 & 100 & 0.74 & 0.00 & 0.00 & 0.00 & 0.50 & 0.00 & 100 & 0.00 \\
11 & D\&M & Alignment & D\&M & 4 & 100 & 1.00 & 0.00 & 0.00 & 0.00 & 1.00 & 0.00 & 100 & 0.00 \\
24 & D\&M & Alignment & D\&M Extensive & 4 & 100 & 0.62 & 0.34 & 0.04 & 0.15 & 0.29 & 0.04 & 100 & 0.00 \\
\hline
\end{tabular}
\caption{Results for Complexity Experiments with Component Modes, Occurrence, Normalized Quality and Entropy Metrics ($Q_{norm}$,
$H_{norm}$), Selective Accuracy (SA), and Completeness and the metrics mean and standard deviations}
\label{tab:extra_experiments}
\end{table*}

The overall results for Input Parameter vs Dataset imply that the model has preferences among the Tilstra HDDSM interaction categories, fails to consistently acknowledge classification decision boundaries embedded in the prompt, and tends to incorporate patterns or system knowledge from pretrained data when systems likely appear in the training set—or even fully defaults to training-data priors when parameter–dataset alignment is imperfect. The resulting behavior ranges from perfect dependency identification, to confident yet deterministically incorrect full-present matrix classifications, to high-variance, low-accuracy outputs, underscoring susceptibility to hallucination-like behavior when semantic constraints are weak or when dependency-type definitions permit broad interpretation.

\subsection{Complexity}
The Complexity experiments compare system decompositions for real systems. We vary (i) subsystem complexity, (ii) writing complexity, and (iii) multi-subsystem hierarchical complexity to evaluate ambiguity/noise, scalability, and hierarchical identification. These experiments provide insight into both component and dependency identification.

\subsubsection{Subsystem comparison}
The drainage \& moisture control subsystem input-parameter–vs-dataset experiments showed inconsistent subsystem dependency identification. Despite these results, the subsystem-complexity experiment was performed to test whether the behavior relates to pretrained knowledge and to semantic reasoning specific to the Drainage \& Moisture Control (D\&M) subsystem. The refrigeration system was selected for its higher decomposition complexity and because it is likely the most common refrigerator subsystem, potentially better represented in LLM training data; this selection enables assessment of whether the model accurately identifies interactions. Based on the ground-truth dataset, a technical document was generated with a manually constructed system decomposition containing symmetric, directly defined dependencies, visualized in \autoref{lst:ManREF.txt}. The best-performing input-parameter–vs-dataset settings from earlier experiments (defined existing system and dependency type) were reused. For the refrigeration system, the input parameter \textit{Liquid Material} was chosen to investigate over-generalization, as it is the most frequent dependency type in the decomposition (22/37, 59\% of direct dependencies). \autoref{tab:extra_experiments} presents results and compares them to Experiment 1.2; performance is notably similar, and the metrics align more closely with Experiment 1.2 than with any of the other experiments. In combination with the Experiment 1.2 findings in \autoref{tab:experiments_full}, this pattern suggests possible over-generalization: in both systems, the most prevalent dependency type performs the worst when present in the dataset. Notably, both dependency types belong to the same HDDSM “\textit{Material}” category, so LLM semantic reasoning may also influence identification.

To assess component and dependency identification under increased decomposition complexity, the Gas Material dependency type was selected because it occurs in both the drainage \& moisture control and refrigeration subsystems, thereby eliminating variability in dependency-type behavior.\autoref{tab:extra_experiments} summarizes overall performance metrics for both subsystems. Although refrigeration appears to perform better overall, examining quality for the directly stated dependencies described in \autoref{lst:ManM&D.txt} and \autoref{lst:ManREF.txt} reveals a different picture: $Q_{D\&M}=1$ with $NZF_{D\&M} = 0.17$, versus $Q_{REF}=0.3$ with $NZF_{D\&M} = 0.09$. Cell-wise comparison indicates perfect classification for the D\&M subsystem and consistent but incorrect classification for the refrigeration system, which falsely marks the dependencies as absent. Considering the NZF of the \textit{Gas Material} dependency for both cases indicates that the Refrigeration subsystem struggles to identify this dependency type, whereas the D\&M subsystem achieves a perfect score. These findings suggest that complexity may affect the identification of the system decomposition: dependency types with lower NZF values may be more difficult for Auto-DSM to identify and should be examined closely when evaluating system-identification performance.
\begin{table*}[h!]
\centering
\renewcommand{\arraystretch}{1.2}
\setlength{\tabcolsep}{5pt}
\begin{tabular}{lllll}
\hline
\textbf{System Parameter}                                                                              & \textbf{REF} & \textbf{D\&M} & \textbf{\begin{tabular}[c]{@{}l@{}}GEN-DSM \\ REF\end{tabular}} & \textbf{\begin{tabular}[c]{@{}l@{}}GEN-DSM\\ D\&M\end{tabular}} \\ \hline
Refrigeration System; Drainage and   Moisture Control System                                           & 7/7          & 0/4           & 7/7                                                             & 0/7                                                             \\
Refrigerator's Refrigeration and Drainage/Moisture Control Subsystems                                  & 7/7          & 1/4           & 7/7                                                             & 0/7                                                             \\
Drainage/Moisture Control and Refrigeration Subsystem                                                  & 7/7          & 4/4           & 7/7                                                             & 0/7                                                             \\
Combined Drainage/Moisture Control and Refrigeration Subsystem of a Refrigerator                       & 7/7          & 0/4           & 7/7                                                             & 0/7                                                             \\
Refrigerator Subsystems: Refrigeration System and Drainage/Moisture Control                            & 7/7          & 1/4           & 7/8                                                             & 1/8                                                             \\
Refrigerator decomposition including both the Refrigeration and Drainage/Moisture Control Subsystem    & 7/7          & 1/4           & 7/8                                                             & 0/8                                                             \\
Refrigerator decomposition including both the Drainage \& Moisture Control and Refrigeration Subsystem & 7/7          & 4/4           & 7/11                                                            & 4/11                                                            \\ \hline
\end{tabular}
\caption{Refrigeration System (REF) and Drainage \& Moisture Control System (D\&M) coverage with Generated DSM (GEN-DSM) component counts}
\label{tab:ref_dsm_no_nzf}
\end{table*} 

\subsubsection{Writing complexity}
Across experiments, a relatively simple technical document was used as input; however, LLM semantic reasoning must be considered. To test the effect of writing complexity on LLM behavior, the synthetically generated technical report in \autoref{lst:D&Mext.txt} was compared to the manually generated document in \autoref{lst:ManM&D.txt}. The synthetic report produces a formal system-decomposition narrative in fluent technical English. To isolate differences, the same input parameters and dependency type were used. Given inconsistency across dependency types, two cases were compared (see \autoref{tab:extra_experiments}): leaving the dependency type empty to expose over-generalization, and using the high-performing \textit{Alignment} type. Whereas Experiment 3.2 was already close to an all-present dependency response, Experiment 22 returned, for $N=30$, thirty fully dependent matrices. Likewise, although Experiment 11 achieved a perfect score, the extensive dataset yielded an almost all-present dependency matrix in which only the dependencies between the moisture control filters and the drain hole toggled between present and absent. \autoref{tab:extra_experiments} shows that—even when the exact dependency type exists in the dataset—sentence-level semantics can re-introduce over-generalization, which must be handled with care. With more extensive datasets, it becomes harder to trace the origin of dependency identification; large documents can obscure incorrect or ambiguous interactions. Future work should add referencing/traceability so engineers can verify whether identifications are based on correct or incorrect text, and incorporate prompt-level structuring constraints to narrow the LLM’s decision space and improve both identification and traceability.

\subsubsection{Multiple system analysis}
Despite limitations revealed by the highest-complexity refrigeration decomposition and by the Input Parameter vs Dataset findings, follow-up experiments with a dual-subsystem decomposition were conducted. A combined dataset was constructed that included both the Refrigeration System and the Drainage \& Moisture Control System, incorporating intra-subsystem interactions and inter-subsystem dependencies (\autoref{lst:ManInter.txt}).

With the system input parameter \textit{Refrigerator}, only the Refrigeration subsystem was returned. Although linguistic similarity could explain this, Experiments 16–19 showed that D\&M components are valid responses even when both subsystems and a dual system definition are present. Consequently, combinations of input parameters were tested. The inputs in \autoref{tab:ref_dsm_no_nzf} examine alternative phrasings for dual-subsystem definitions. The resulting GEN-DSMs consistently included only Refrigeration components; identification stabilized on a single subsystem depending on wording.

Sensitivity to system-parameter phrasing was evaluated across six structured variations (\autoref{tab:ref_dsm_no_nzf}). In all but one case, outputs contained complete Refrigeration sets (7/7) and at most one drainage-related component. Output logs indicated that Auto-DSM extracted components for both subsystems but stored them as two lists, after which only the first list was used for dependency-type identification. A notable exception was the phrasing \textit{“Refrigerator decomposition including both the Drainage \& Moisture Control and Refrigeration Subsystem”}, which yielded a single list; in 21/30 runs, all 11 components—7 Refrigeration and 4 D\&M—were included in the GEN-DSM.

While this phrasing demonstrated that the LLM is capable of jointly recognizing both subsystems, it shows the effect of poorly constrained prompts
Where just only defining the input parameter differently allows for such different outcomes. Further analysis revealed that the order in which subsystems are mentioned in the input parameter directly impacts the outcome.
This asymmetry provides strong evidence that the LLM processes composite noun phrases in a sequential manner, frequently prioritizing the first identified entity. It appears that only one system listing is actively used to construct the DSM, even if there are two acknowledged during textual reasoning. In conclusion, although semantically refined input phrasing improves component identification, the model remains unable to robustly construct structured DSMs involving multiple interacting subsystems. The results suggest that further intervention—either in the form of prompt decomposition, architectural support for subsystem segmentation, or DSM post-processing—is necessary to ensure consistency, modularity, and dependency accuracy in multiple system decomposition tasks.

These results are first-order observations: Auto-DSM degrades with larger decompositions and multi-subsystem scope, exhibiting over-generalization, first-entity bias, and weak boundary acknowledgment, while aggregate metrics may mask cell-level errors rooted in pretrained patterns. Differences across complexity factors—including writing style, subsystem structure, and multi-system hierarchy—should be carefully examined to clarify overall behavior and to guide strategies for reducing errors through stricter prompt design, subsystem-aware processing, and text-to-dependency traceability.

%% file: 6_Discussion/Discussion.tex
\section{Discussion}

\subsection{Evaluation Boundaries and Dataset Circularity}
The use of ChatGPT to construct both the ground-truth dataset and the input documentation introduces a circular evaluation risk. By evaluating Auto-DSM on documents structurally aligned with its own pretraining distribution, performance may reflect pattern recognition rather than genuine system reasoning. While this approach offers control for a first-order evaluation, future research must extend the framework to domain-specific, out-of-distribution systems. Testing with proprietary industrial data would allow clearer distinction between memorization and generalizable decomposition capabilities.

\subsection{Reasoning Under Complexity and Context Scarcity}
The complexity experiments highlight Auto-DSM’s limitations in handling ambiguous inputs, multi-subsystem hierarchies, and low-context prompts. While component identification remained robust, dependency identification degraded as document length and structural depth increased. These results align with findings from Shi et al. \cite{shi2024trusting}, suggesting that in the absence of strong context, LLMs revert to prior training distributions. This behavior is especially evident in overgeneralization, hallucinated links, and first-entity bias, indicating that increased prompt clarity alone is insufficient without architectural support for contextual integration.

\subsection{Prompt Sensitivity and Interaction Definitions}
Auto-DSM’s outputs are highly sensitive to both the formulation and consistency of input definitions. Bi-directional phrasing improved symmetry and reduced entropy, but confidently incorrect classifications persisted, especially for undefined or contradictorily defined interactions. These effects reveal a lack of internal checks for semantic consistency, and highlight the necessity of standardized interaction schemas within the pipeline. Furthermore, when definitions are redundant or conflicting, the model’s outputs vary depending on the order and phrasing of input, revealing its sensitivity to sequence effects and susceptibility to prompt-based bias. This highlights the need for stricter control over input parameter formatting.

\subsection{Current Pipeline Limitations}
Despite promising initial results, the Auto-DSM pipeline is not yet robust for deployment in complex or safety-critical engineering contexts. Major limitations include:
\begin{itemize}
    \item Limited presence of hallucination detection or abstention mechanisms.
    \item No support for multi-layered or hierarchical system structures.
    \item High sensitivity to prompt wording and insufficient robustness to low-context scenarios.
    \item No internal error handling or validation of malformed inputs.
    \item Lack of dependency traceability prevents verification of identified interactions, making it difficult to trace hallucinated or incorrect links back to their source in the input document. 
\end{itemize}

The evaluation focused primarily on dependency identification. Although component identification remained stable across experiments, future use cases may demand extended metrics to assess component recognition, especially in larger systems or those with non-standard naming conventions.

\subsection{Beyond DSM: Toward Flexible Representations}
Current DSM representations, while structured and interpretable, may not fully capture the goal of the decomposition. Future work could explore integrating Auto-DSM outputs into executable system representations, such as the Elephant Specification Language (ESL) or constraint-based models. This could support richer downstream tasks, including simulation, validation, and automated design-space exploration.

\subsection{Pathways for Improvement}
Refinements to the current framework should include:
\begin{itemize}
    \item Schema-constrained prompt engineering with fixed dependency categories.
    \item Subsystem-aware architecture to support multi-layer decomposition.
    \item Retrieval-augmented generation or agent-based verification to reduce hallucination.
    \item Evaluation on unseen, real-world datasets to measure generalization performance.
    \item Investigation of alternative linguistic formulations for dependencies (e.g., “depends on”, “interacts with”) to assess model robustness to phrasing variations.
\end{itemize}

These directions will contribute to a more interpretable, accurate, and scalable system decomposition methodology that bridges data-driven reasoning with engineering traceability.

%% file: 7_Conclusion_Future_work/Conclusion.tex
\section{Conclusion}
This paper presented a black-box evaluation framework for assessing LLM-based system decomposition in the context of Design Structure Matrices (DSM). The framework was designed to be model-agnostic, robust to evolving system complexity, and capable of capturing both structural accuracy and behavioral patterns in LLM outputs. It introduced metrics for correctness, consistency, abstention, and entropy—providing a comprehensive lens for evaluating performance beyond raw accuracy.

To address the central research question, the framework enables evaluation of decomposition quality, dependency classification, and abstention behavior using controlled, ground-truth datasets. Repeated experiments across prompt formulations, dependency definitions, and complexity levels exposed several behavioral patterns: overconfidence in hallucinated links, sensitivity to input phrasing, first-entity bias in composite system descriptions, and semantic inconsistency under complexity. These outcomes confirm that even within known systems, Auto-DSM struggles to generalize robustly, especially when semantic clarity or contextual cues are weak.

While the current evaluation leverages training-aligned documents for initial validation, future extensions must target systems unfamiliar to the model to test true generalization. Moreover, the findings motivate several architectural and methodological refinements, including improved prompt schema design, multi-system input handling, and integration with more expressive representations such as ESL or constraint models.

Ultimately, this paper provides a foundational evaluation strategy for automated system decomposition and a roadmap for advancing LLM-enabled engineering tools within the DSM domain. As system complexity increases and models evolve, such frameworks will be essential for ensuring traceable, verifiable, and reliable integration of AI into engineering design processes.

%% file: A_Appendices/Appendix_A.tex
\newpage
\appendices
\onecolumn

\section{Current situation Auto-DSM}\label{Ap:Datasets}

{\begin{table}[ht]
\centering
\caption{Occurrence Frequency (\%) Over 10 Runs of Auto‑DSM (\textit{Engine \& Mechanical}; Dataset \cite{Lakshminarayanan2020})}
\label{tab:frequency}
\begin{tabular}{|l|llllll|}
\hline
\textbf{Frequency} & Cylinder head                                       & Crankcase                                           & Connecting rod                                      & Pistons & Camshaft                                            & Crankshaft                                          \\ \hline
Cylinder head      & 100\%                                               & \cellcolor[HTML]{FFEB9C}{\color[HTML]{9C5700} 90\%} & 100\%                                               & 100\%   & 100\%                                               & 100\%                                               \\
Crankcase          & 100\%                                               & 100\%                                               & \cellcolor[HTML]{FFC7CE}{\color[HTML]{9C0006} 70\%} & 100\%   & \cellcolor[HTML]{C6EFCE}{\color[HTML]{006100} 90\%} & \cellcolor[HTML]{C6EFCE}{\color[HTML]{006100} 90\%} \\
Connecting rod     & \cellcolor[HTML]{FFEB9C}{\color[HTML]{9C5700} 90\%} & \cellcolor[HTML]{FFEB9C}{\color[HTML]{9C5700} 90\%} & 100\%                                               & 100\%   & 100\%                                               & 100\%                                               \\
Pistons            & 100\%                                               & 100\%                                               & 100\%                                               & 100\%   & \cellcolor[HTML]{FFC7CE}{\color[HTML]{9C0006} 60\%} & 100\%                                               \\
Camshaft           & \cellcolor[HTML]{FFEB9C}{\color[HTML]{9C5700} 90\%} & 100\%                                               & 100\%                                               & 100\%   & 100\%                                               & 100\%                                               \\
Crankshaft         & \cellcolor[HTML]{FFC7CE}{\color[HTML]{9C0006} 70\%} & 100\%                                               & 100\%                                               & 100\%   & 100\%                                               & 100\%                                               \\ \hline
\end{tabular}
\end{table}
}

\begin{table}[ht]
\centering
\caption{Most Occurring Response Over 10 Runs of Auto‑DSM (\textit{Engine \& Mechanical}; Dataset \cite{Lakshminarayanan2020})}
\label{tab:most-occurring}
\begin{tabular}{|l|llllll|}
\hline
\textbf{Most occurring value} & Cylinder head                                    & Crankcase                                        & Connecting rod                                   & Pistons & Camshaft                                         & Crankshaft                                       \\ \hline
Cylinder head                 & 1                                                & \cellcolor[HTML]{FFEB9C}{\color[HTML]{9C5700} 1} & 0                                                & 1       & 1                                                & 0                                                \\
Crankcase                     & 1                                                & 1                                                & \cellcolor[HTML]{FFC7CE}{\color[HTML]{9C0006} 1} & 0       & \cellcolor[HTML]{C6EFCE}{\color[HTML]{006100} 5} & \cellcolor[HTML]{C6EFCE}{\color[HTML]{006100} 1} \\
Connecting rod                & \cellcolor[HTML]{FFEB9C}{\color[HTML]{9C5700} 0} & \cellcolor[HTML]{FFEB9C}{\color[HTML]{9C5700} 1} & 1                                                & 1       & 0                                                & 1                                                \\
Pistons                       & 1                                                & 1                                                & 1                                                & 1       & \cellcolor[HTML]{FFC7CE}{\color[HTML]{9C0006} 1} & 1                                                \\
Camshaft                      & \cellcolor[HTML]{FFEB9C}{\color[HTML]{9C5700} 1} & 0                                                & 0                                                & 0       & 1                                                & 1                                                \\
Crankshaft                    & \cellcolor[HTML]{FFC7CE}{\color[HTML]{9C0006} 1} & 1                                                & 1                                                & 1       & 1                                                & 1                                                \\ \hline
\end{tabular}
\end{table}

\begin{table}[ht]
\centering
\caption{Number of Rejection Responses Over 10 Runs of Auto‑DSM (\textit{Engine \& Mechanical}; Dataset \cite{Lakshminarayanan2020})}
\label{tab:idk-counts}
\begin{tabular}{|l|llllll|}
\hline
\textbf{Rejection} & Cylinder head                                    & Crankcase                                        & Connecting rod                                   & Pistons & Camshaft                                         & Crankshaft                                       \\ \hline
Cylinder head      & 0                                                & \cellcolor[HTML]{FFEB9C}{\color[HTML]{9C5700} 0} & 0                                                & 0       & 0                                                & 0                                                \\
Crankcase          & 0                                                & 0                                                & \cellcolor[HTML]{FFC7CE}{\color[HTML]{9C0006} 0} & 0       & \cellcolor[HTML]{C6EFCE}{\color[HTML]{006100} 9} & \cellcolor[HTML]{C6EFCE}{\color[HTML]{006100} 1} \\
Connecting rod     & \cellcolor[HTML]{FFEB9C}{\color[HTML]{9C5700} 0} & \cellcolor[HTML]{FFEB9C}{\color[HTML]{9C5700} 0} & 0                                                & 0       & 0                                                & 0                                                \\
Pistons            & 0                                                & 0                                                & 0                                                & 0       & \cellcolor[HTML]{FFC7CE}{\color[HTML]{9C0006} 0} & 0                                                \\
Camshaft           & \cellcolor[HTML]{FFEB9C}{\color[HTML]{9C5700} 0} & 0                                                & 0                                                & 0       & 0                                                & 0                                                \\
Crankshaft         & \cellcolor[HTML]{FFC7CE}{\color[HTML]{9C0006} 0} & 0                                                & 0                                                & 0       & 0                                                & 0                                                \\ \hline
\end{tabular}
\end{table}


\twocolumn

\section{Prompt Engineering} \label{sec: Appendix A}
\begin{center}
\begin{promptbox}{Prompt 1 - Define System Structure}
\# Role:\\
You are a systems engineer specialized in technical system decomposition and dependency analysis.\\

\# Task:\\
Generate a complete system decomposition of a standard household refrigerator, existing of subsystems and components.\\

\# Instructions:\\
- Identify all main subsystems.\\
- For each subsystem, list its key components.\\
- Use only realistic and commonly found components.\\

\# Constraints:\\
- Do not invent components or subsystems.\\
- Use consistent formatting across all subsystems.\\

\# Output Format:\\
\noindent{\ttfamily [Subsystem Name 1]}\\
\noindent{\ttfamily • [Component A]}\\
\noindent{\ttfamily • [Component B]}\\
...\\
\noindent{\ttfamily [Subsystem Name 2]}\\
\noindent{\ttfamily • [Component F]}\\
\noindent{\ttfamily • [Component G]}\\
...\\
\end{promptbox}
\captionof{figure}{Prompt 1 used to define the refrigerator subsystem decomposition and its components.}
\label{lst:prompt1}
\end{center}

\begin{center}
\begin{promptbox}{Prompt 2 - Intra-Subsystem Dependency Mapping}
\# Role:\\
You are a systems engineer specialized in technical system decomposition and dependency analysis.\\

\# Task:\\
Identify all intra-subsystem dependencies between components in a single subsystem.\\

\# Instructions:\\
- For each component pair within the same subsystem, determine whether a dependency exists.\\
- If yes, classify it using only Tilstra’s (2012) HDDSM interaction types.\\

\# Constraints:\\
- Do not invent components or interaction types.\\
- Complete one subsystem at a time.\\
- Output must follow the structured tabular format.\\

\# Interaction Types (Tilstra, 2012):\\
1. Information:    \hspace*{2em}1.1 Status [SI], 1.2 Control [CI]\\
2. Material:
    \hspace*{2em}2.1 Human [HM], 2.2 Gas [GM], 2.3 Liquid [LM], 2.4 Solid [SM], 2.5 Plasma [PM], 2.6 Mixture [MM]\\
3. Energy:
   \hspace*{2em}3.1 Human [HE], 3.2 Acoustic [AE], 3.3 Biological [BE], 3.4 Chemical [CE], 3.5 Electrical [EE], 
   3.6 Electromagnetic [EME], 3.7 Hydraulic [HYE], 3.8 Mechanical [ME], 3.9 Magnetic [MAG], 
   3.10 Pneumatic [PE], 3.11 Radioactive [NE], 3.12 Thermal [TE], 3.13 Strain energy [SE]\\
4. Spatial:
   \hspace*{2em}4.1 Proximity [P], 4.2 Alignment [A]\\
5. Movement:
   \hspace*{2em}5.1 Translational [LRM], 5.2 Rotational [RRM]\\

\# Subsystem and its decomposition:\\
\{Insert Subsystem + Components from output prompt 1\}\\

\# Output Format:\\
\noindent{\ttfamily [Subsystem;	From (Component);	To (Component);	Interaction Type(s);	...;	Target Subsystem]}\\

\# Example Template:\\
\noindent{\ttfamily Subsystem 1;	Component A;	Component B;	Interaction Type 1 [Abbreviation];	Interaction Type 2 [Abbreviation]; ...;	Subsystem 1}\\
\end{promptbox}
\captionof{figure}{Prompt 2 for intra-subsystem interaction identification using Tilstra (2012) HDDSM.}
\label{lst:prompt2}
\end{center}

\begin{center}
\begin{promptbox}{Prompt 3 - Inter-Subsystem Dependency Mapping (Optimized Version)}
\# Role:\\
You are a systems engineer specialized in technical system decomposition and dependency analysis.\\

\# Task:\\
Identify all inter-subsystem dependencies between components of the specified Subsystem and the remaining Subsystems.\\

\# Specified Subsystem and its decomposition:\\
Name: \{Insert Subsystem A + Components from prompt 1\} \\

\# Remaining Subsystems:\\
\{Insert remaining Subsystems + Components from prompt 1\} (previously defined)\\

\# Instructions:\\
- Compare each component from the specified Subsystem to all the components of the remaining Subsystems.\\
- Use **only the components defined** in the reamining subsystem list for the dependency analysis.\\
- If a valid dependency exists, classify it using the interaction types from Tilstra’s (2012) HDDSM framework.\\

\# Constraints:\\
- Do **not invent** any components or interaction types.\\
- Do **not assume** new relationships. Only use information available in the subsystem decomposition.\\
- Follow the output format precisely.\\

\# Output Format:\\
\noindent{\ttfamily [Subsystem; From (Component); To (Component); Interaction Type(s); ...; Target Subsystem]}\\

\# Example Template:\\
\noindent{\ttfamily  Subsystem 1;	Component A;	Component G;	Interaction Type 1 [Abbreviation];	Interaction Type 2 [Abbreviation]; ...;	Subsystem 2}\\
\end{promptbox}
\captionof{figure}{Prompt 3 for inter-subsystem dependency mapping using explicit injection of a specified Subsystem and controlled reference to the remaining subsystems.}
\label{lst:prompt3}
\end{center}

\begin{center}
    \begin{promptbox}{Prompt 1 - System Decomposition Article Generation (Excel-based)}
\# Role:\\
You are a mechanical systems engineer and technical writer, specialized in decomposing systems. Your expertise lies in translating structured technical data into formal, traceable system decomposition reports for engineering documentation. \\

\# Task:\\
Write a clear, technically accurate, traceable and well-structured system decomposition article for the subsystem $**${SUBSYSTEM\_NAME}$**$, which is part of the system $**$Refrigerator$**$. Use the structured Excel data as source for the system decomposition article. This Excel file contains one row per interaction, formatted according to the structure defined below.\\

\# Data Format:\\
Each row contains:\\
1. [INDEX] — Unique ID for the interaction  \\
2. [Subsystem] — Subsystem of the source component (Component A) \\ 
3. [From (Component)] — Source component (Component A) \\
4. [To (Component)] — Target component  (Component B)\\
5–7. [Interaction Type(s)] — One or more interaction types (Tilstra classification)  \\
8. [Target Subsystem] — Subsystem of the target component (Component B)\\

Use the following interaction classification when interpreting types:\\
1. Information\\
- [SI] Status\\
- [CI] Control\\
2. Material\\
- [HM] Human\\
- [GM] Gas\\
- [LM] Liquid\\
- [SM] Solid\\
- [PM] Plasma\\
- [MM] Mixture\\
3. Energy\\
- [HE] Human\\
- [AE] Acoustic\\
- [BE] Biological\\
- [CE] Chemical\\
- [EE] Electrical\\
- [EME] Electromagnetic\\
- [HYE] Hydraulic\\
- [ME] Mechanical\\
- [MAG] Magnetic\\
- [PE] Pneumatic\\
- [NE] Radioactive\\
- [TE] Thermal\\
- [SE] Strain energy\\
4. Spatial\\
- [P] Proximity\\
- [A] Alignment\\
5. Movement\\
- [LRM] Translational\\
- [RRM] Rotational\\
When referring to an interaction, include both its name + category and abbreviation, e.g., Thermal Energy [TE].\\

\# Instructions:\\
The article should:\\
1. Begin with a concise, high-level introduction explaining the role and function of `{SUBSYSTEM\_NAME}` within the full system `{SYSTEM}`.\\
2. Present all interactions in a smooth, technical narrative.\\
3. Discuss each interaction once, referring to its unique `[INDEX]` (e.g., "[5]") for traceability.\\
4. Indicate whether each interaction is intra-subsystem (within the same subsystem) or inter-subsystem (across subsystems).\\
5. For each interaction:\\
   - Name the source and target components.\\
   - Describe all interaction types using their **Tilstra classification**.\\
   - Group related components or functions logically to enhance readability and flow.\\
6. Write in fluent, formal, technical English, consistent with engineering documentation standards.\\

\# Constraints:\\
- **No new components, interactions, or interaction types** may be added, inferred, or assumed beyond the input data.\\
- Use **every input row exactly once**, no skipping or duplication.\\
- Only use interaction types from the **Tilstra (2012)** classification. Do not modify, rename, or combine categories.\\
- Always include both the **full name and abbreviation** for interaction types, e.g., *Thermal Energy [TE]*.\\
- Avoid using bullet points or lists in the article unless strictly necessary — maintain a **narrative format**.\\
- Do not generalize or invent functionality that is not explicitly described in the data.\\

\# Example:\\
**Excel Input Row:**\\
INDEX: 1; Subsystem: Refrigeration System; From: Compressor; To: Condenser coil; Interaction Types: Liquid [LM], Thermal [TE], Mechanical [ME]; Target Subsystem: Refrigeration System\\

**Correct Response:** \\
The compressor mechanically and thermally interacts with the condenser coil, delivering pressurized refrigerant in liquid form [1]. As both components belong to the Refrigeration System, this is classified as an intra-subsystem interaction.\\

\# Input:
Here is the input table, filtered for \{SUBSYSTEM\_NAME\}:\\
{Chat\_input.xlsx}\\
Use this data to generate the full system decomposition article as described above.\\
\end{promptbox}
\captionof{figure}{Prompt for generating a system decomposition article using structured interaction data and Tilstra's (2012) classification framework.}
\label{lst:experprompt1}
\end{center}

\vspace{1em}

\clearpage
\onecolumn

\section{Datasets}\label{Ap:Datasets}
\begin{figure}[H]
    \centering
    \includegraphics[width=0.5\linewidth]{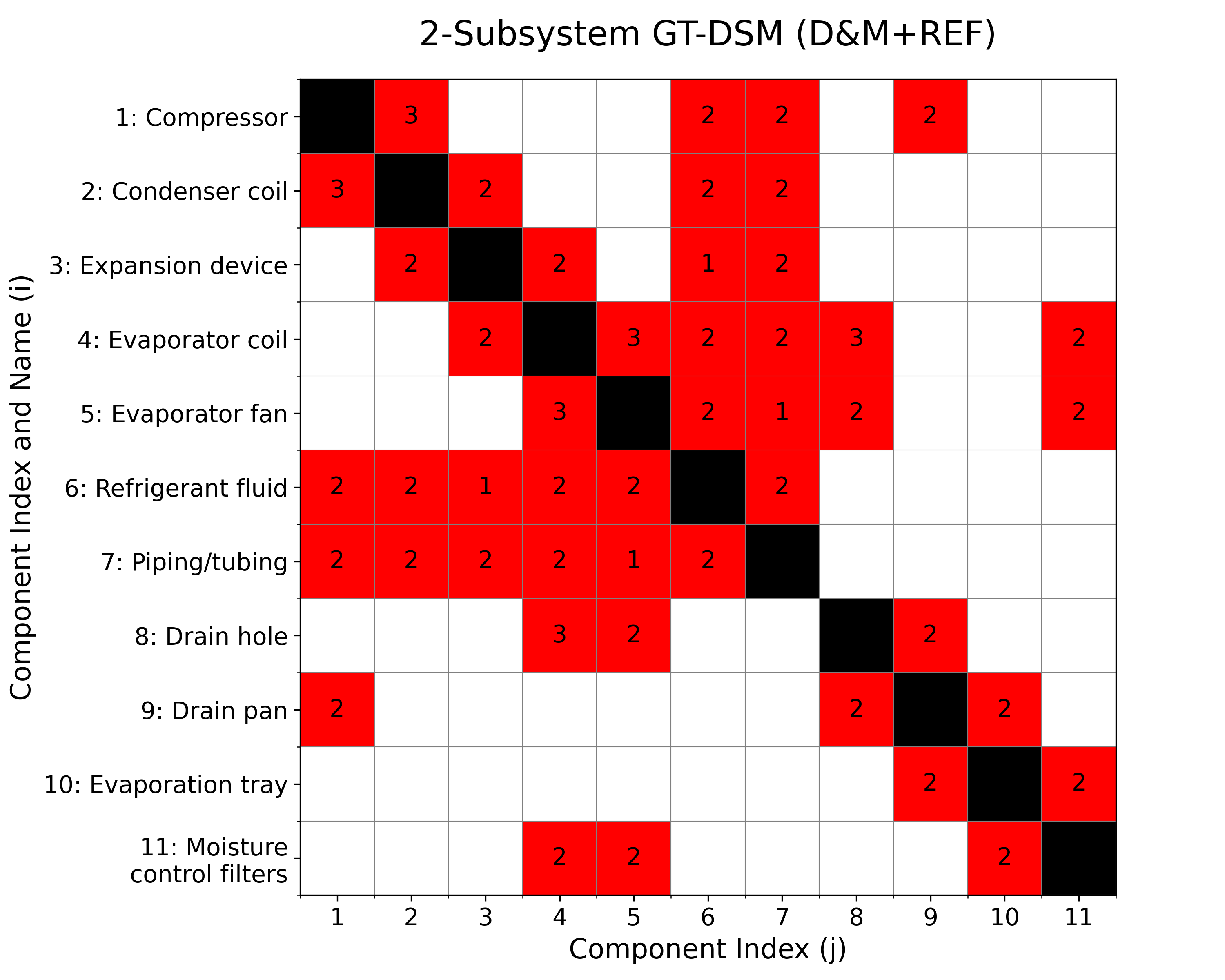}
    \caption{GT-DSM for the Two Combined Subsystems: Drainage \& Moisture Control System and Refrigeration System}
    \label{fig:GT-DSM:REF+D&M}
\end{figure}

{
\scriptsize
\rowcolors{2}{lightgray}{white}
\setlength{\tabcolsep}{4pt}
\begin{table}[htbp]
\centering
\begin{tabularx}{\textwidth}{|X|X|X|X|X|X|X|X|}
\rowcolor{headerblue}
\hline
\textbf{Index} & \textbf{Subsystem} & \textbf{From (Component)} & \textbf{To (Component)} & \textbf{Interaction Type 1} & \textbf{Interaction Type 2} & \textbf{Interaction Type 3} & \textbf{Target Subsystem} \\
\hline
1 & Refrigeration System & Compressor & Condenser coil & 2.3 Liquid [LM] & 3.12 Thermal [TE] & 3.8 Mechanical [ME] & Refrigeration System \\
2 & Refrigeration System & Compressor & Refrigerant fluid & 2.3 Liquid [LM] & 3.8 Mechanical [ME] & & Refrigeration System \\
3 & Refrigeration System & Compressor & Piping / tubing & 4.1 Proximity [P] & 3.8 Mechanical [ME] & & Refrigeration System \\
4 & Refrigeration System & Compressor & Microcontroller / control board & 1.2 Control [CI] & 3.5 Electrical [EE] & & Control System \\
5 & Refrigeration System & Compressor & Thermostat or temperature sensors & 1.1 Status [SI] & & & Control System \\
6 & Refrigeration System & Compressor & Relay switches & 3.5 Electrical [EE] & & & Control System \\
7 & Refrigeration System & Compressor & Power supply \& electrical wiring & 3.5 Electrical [EE] & & & Control System \\
8 & Refrigeration System & Compressor & Outer cabinet & 3.8 Mechanical [ME] & 4.1 Proximity [P] & & Structural \& Insulation System \\
9 & Refrigeration System & Compressor & Drain pan & 3.12 Thermal [TE] & 4.1 Proximity [P] & & Drainage \& Moisture Control System \\
10 & Refrigeration System & Compressor & Control panel (buttons/display) & 1.2 Control [CI] & & & User Interface \& Access System \\
11 & Refrigeration System & Compressor & Defroster timer / control board & 1.2 Control [CI] & & & Defrost System \\
12 & Refrigeration System & Condenser coil & Expansion device & 2.3 Liquid [LM] & 3.12 Thermal [TE] & & Refrigeration System \\
13 & Refrigeration System & Condenser coil & Refrigerant fluid & 2.3 Liquid [LM] & 3.12 Thermal [TE] & & Refrigeration System \\
14 & Refrigeration System & Condenser coil & Piping / tubing & 4.1 Proximity [P] & 2.3 Liquid [LM] & & Refrigeration System \\
15 & Refrigeration System & Condenser coil & Outer cabinet & 4.1 Proximity [P] & & & Structural \& Insulation System \\
16 & Refrigeration System & Condenser coil & Compressor & 2.3 Liquid [LM] & 3.12 Thermal [TE] & 3.8 Mechanical [ME] & Refrigeration System \\
17 & User Interface \& Access System & Control panel (buttons/display) & Door(s) & 4.1 Proximity [P] & & & User Interface \& Access System \\
18 & User Interface \& Access System & Control panel (buttons/display) & Microcontroller / control board & 1.1 Status [SI] & 1.2 Control [CI] & 3.5 Electrical [EE] & Control System \\
19 & User Interface \& Access System & Control panel (buttons/display) & Power supply \& electrical wiring & 3.5 Electrical [EE] & & & Control System \\
20 & User Interface \& Access System & Control panel (buttons/display) & Lighting circuit & 1.2 Control [CI] & & & Lighting System \\
21 & User Interface \& Access System & Control panel (buttons/display) & Light switch / sensor & 1.2 Control [CI] & 3.5 Electrical [EE] & & Lighting System \\
22 & User Interface \& Access System & Control panel (buttons/display) & Compressor & 1.2 Control [CI] & & & Refrigeration System \\
23 & User Interface \& Access System & Control panel (buttons/display) & Expansion device & 1.2 Control [CI] & & & Refrigeration System \\
24 & User Interface \& Access System & Control panel (buttons/display) & Defrost heater & 1.2 Control [CI] & & & Defrost System \\
\hline
\end{tabularx}
\caption{Section of Refrigerator system decomposition dataset}
\label{tab:excel_style_refrigerator}
\end{table}
}
\twocolumn
\normalsize

\clearpage
\onecolumn
\begin{table*}[htbp]
\centering
\scriptsize
\setlength{\tabcolsep}{3pt}
\renewcommand{\arraystretch}{1.2}
\begin{tabular}{|l|l|*{7}{c|}}
\hline
\multirow{2}{*}{\textbf{Category}} & \multirow{2}{*}{\textbf{Metric}} 
& \textbf{Refrigeration} & \textbf{Control} & \textbf{User Interface} & \textbf{Structural \& Insulation} & \textbf{Defrost} & \textbf{Lighting} & \textbf{Drainage \& Moisture} \\
& & \textbf{System} & \textbf{System} & \textbf{Access System} & \textbf{System} & \textbf{System} & \textbf{System} & \textbf{Control System} \\
\hline
\multirow{2}{*}{Subsystem Size} 
& Components & 7 & 4 & 4 & 5 & 4 & 3 & 4 \\
& Entries (Non-diagonal) & 42 & 12 & 12 & 20 & 12 & 6 & 12 \\
\hline
\multirow{3}{*}{\shortstack{Intra-subsystem\\Interactions}}
& Active links & 30 & 12 & 6 & 14 & 10 & 6 & 6 \\
& Inactive links & 12 & 0 & 6 & 6 & 2 & 0 & 6 \\
& NZF (\%) & 71.43 & 100 & 50 & 70 & 83.33 & 100 & 50 \\
\hline
\multirow{3}{*}{\shortstack{Inter-subsystem\\Interactions}}
& Active links & 38 & 32 & 22 & 30 & 16 & 16 & 16 \\
& Inactive links & 850 & 886 & 896 & 880 & 902 & 908 & 902 \\
& NZF (\%) & 4.28 & 3.49 & 2.40 & 3.30 & 1.74 & 1.73 & 1.74 \\
\hline
\multirow{5}{*}{Total} 
& Active links & 68 & 44 & 28 & 44 & 26 & 22 & 22 \\
& Inactive links & 862 & 886 & 902 & 886 & 904 & 908 & 908 \\
& Total \# Interactions & 128 & 60 & 53 & 90 & 47 & 43 & 47 \\
& NZF (\%) & 7.31 & 4.73 & 3.01 & 4.73 & 2.80 & 2.37 & 2.37 \\
& Interaction types & 10 & 4 & 9 & 9 & 6 & 7 & 7 \\
\hline
\end{tabular}
\caption{Subsystem metrics for the refrigerator including component sizes, interaction counts, and NZF values per subsystem}
\label{tab:subsystem_metrics}
\end{table*}
\clearpage
\twocolumn

\section{Technical Documentation}
\begin{lstlisting}[style=coltxt, caption={Manual Drainage \& Moisture Control System Decomposition},label={lst:ManM&D.txt}]
The Drainage & Moisture Control System is composed of components drain hole, drain pan, evaporation tray, and moisture control filters. The components interact to ensure system functionality, as detailed below:

The drain hole is both Mixture Material [MM] and Alignment [A] linked to the drain pan [77]. The drain pan is both Mixture Material [MM] and Alignment [A] linked to the drain hole [83].

The drain pan is both Mixture Material [MM] and Proximity [P] linked to the evaporation tray [84]. The evaporation tray is both Mixture Material [MM] and Proximity [P] linked to the drain pan [89].

The evaporation tray is both Mixture Material [MM] and Gas Material [GM] linked to the moisture control filters [90]. The moisture control filters are both Mixture Material [MM] and Gas Material [GM] linked to the evaporation tray [189].
\end{lstlisting}

\begin{lstlisting}[style=coltxt,caption={Manual Refrigeration System System Decomposition},label={lst:ManREF.txt}]
The Refrigeration System is composed of the compressor, condenser coil, evaporator coil, expansion device, piping/tubing, refrigerant fluid, and evaporator fan. The components interact to ensure system functionality, as detailed below:
    
The compressor is Liquid Material [LM], Thermal Energy [TE] and Mechanical Energy [ME] linked to the condenser coil [1], and is both Liquid Material [LM] and Mechanical Energy linked to the refrigerant fluid [2]. Additionally, the compressor is Proximity [P] and Mechanical
Energy [ME] linked to the piping/tubing [3].

The condenser coil is Liquid Material [LM] and Thermal Energy [TE] linked to the expansiondevice [12] and refrigerant fluid [13]. The condenser coil is also Proximity [P] and Liquid
Material [LM] linked to the piping/tubing [14]. Next to that, the condenser coil is Liquid Material [LM], Thermal Energy [TE] and Mechanical Energy [ME] linked to the compressor [16].

The evaporator coil is Liquid Material [LM] and Thermal Energy [TE] linked to the expansion device [94] and the refrigerant fluid [95]. The evaporator coil is Proximity [P] and Liquid Material [LM] linked to the piping/tubing [96]. The evaporator coil is also Proximity [P], Gas Material [GM], and Thermal Energy [TE] linked to the evaporator fan [106].

Simultaneously, the evaporator fan is Proximity [P], Gas Material [GM], and Thermal Energy [TE] linked to the evaporator coil [107]. The evaporator fan is Proximity [P] linked to the piping/tubing [108], while Proximity [P] and Gas Material [GM] linked to the refrigerant fluid
[109].

The expansion device is Liquid Material [LM] and Thermal Energy [TE] linked to both the condenser coil [120] and evaporator coil [121], and only Liquid Material [LM] linked to refrigerant fluid [122]. It also is linked with the piping/tubing through Proximity [P] and Liquid Material [LM] [123].

The piping/tubing is Proximity [P] and Mechanical Energy [ME] linked to the compressor [202], and Proximity [P] to the evaporator fan [206]. The piping/tubing is Proximity [P] and Liquid Material [LM] linked to the condenser coil [203], expansion device [204], evaporator coil [205], and refrigerant fluid [207].

The refrigerant fluid is Thermal Energy [TE] and Liquid Material [LM] linked to the condenser coil [220], and evaporator coil [222]. It is Liquid Material [LM] linked to the expansion device via [221], and Liquid Material [LM] and Proximity [P] linked to the piping/tubing [223]. The refrigerant fluid is Liquid Material [LM] and Mechanical Energy [ME] linked to the compressor [227], and is Proximity [P] and Gas Material [GM] linked to evaporator fan [228].
\end{lstlisting}

\begin{lstlisting}[style=coltxt,caption={Manual Inter-subsystem Refrigeration and Drainage \& Moisture Control System Decomposition},label={lst:ManInter.txt}]
The refrigeration subsystem interacts with the Drainage & Moisture Control System, and all interactions described below occur between these two subsystems, making them intersubsystem.

The compressor is Thermal Energy [TE] and Proximity [P] linked to the drain pan [9]. The drain pan is also Thermal Energy [TE] and Proximity [P] linked to the compressor [86].

The evaporator coil is Proximity [P], Mixture Material [MM] and Liquid Material [LM] linked to the drain hole [101]. The drain hole also is Proximity [P], Mixture Material [MM] and Liquid Material [LM] linked to the evaporator coil [79]. 

The drain hole is Gas Material [GM] and Thermal Energy [TE] linked to the evaporator fan [78]. The evaporator fan, in turn, is Gas Material [GM] and Thermal Energy [TE] linked to the drain hole [118].

The evaporator fan is Gas Material [GM] and Mixture Material [MM] linked to the moisture control filters [119]. Similarly, the evaporator coil is Gas Material [GM] and Mixture Material [MM] linked to the moisture control filters [105]. 

Lastly, the moisture control filters are Gas Material [GM] and Mixture Material [MM] linked to the evaporator fan [185]. They are also Gas Material [GM] and Mixture Material [MM] linked to the evaporator coil [188].
\end{lstlisting}

\begin{lstlisting}[style=coltxt, caption={Generated Technical Narritive Document Drainage \& Moisture Control System Decomposition},label={lst:D&Mext.txt}]
System Decomposition: Drainage & Moisture Control System (Refrigerator)

The Drainage & Moisture Control System in the refrigerator ensures that excess water and moisture are effectively collected, transferred, evaporated, and managed. Its primary function is to keep the interior dry and free of condensation, which helps prevent ice buildup, odors, and mold formation. This subsystem consists of four key components: the drain hole, drain pan, evaporation tray, and moisture control filters.

The process begins with the drain hole, which directs melted water into the drain pan. This interaction involves the movement of a Mixture [MM]—typically water with minor impurities—and is guided by Alignment [A] to ensure proper flow between components [77]. Since both elements belong to the same subsystem, this is an intra-subsystem interaction.

In return, the drain pan also connects back to the drain hole, reinforcing the same pathway. It receives the water mixture, where spatial alignment plays a key role in maintaining the correct flow direction. This interaction again involves Mixture [MM] and Alignment [A] and remains within the same subsystem [83].

Next, the drain pan sends collected water to the evaporation tray. This is a straightforward transfer of a Mixture [MM] aided by Proximity [P], allowing the water to flow passively between the two components [84]. This too is intra-subsystem.

The evaporation tray can also return water to the drain pan, for instance during redistribution or overflow. As with the previous interaction, the evaporation tray and drain pan interaction involves a Mixture [MM] and relies on Proximity [P] to allow the liquid to move freely between both surfaces [89]. The evaporation tray and the moisture control filters are closely coupled in managing moisture release and reabsorption. As water in the evaporation tray turns to vapor, the resulting moisture-rich air flows toward the filters. This process involves both Mixture [MM] and Gas [GM], capturing the physical phase change and directional vapor transfer from tray to filters [90]. Conversely, the moisture control filters also influence the evaporation environment by releasing filtered or redirected moisture-laden air back toward the evaporation tray. This return interaction likewise involves Gas [GM] and Mixture [MM], reinforcing the ambient humidity exchange near the tray [189]. Both are intra-subsystem interactions and together form a bidirectional moisture regulation mechanism that supports efficient and consistent evaporation.
\end{lstlisting}